\documentclass[12pt]{article}

\usepackage{times}
\usepackage{epsfig}
\usepackage{graphicx,color}
\usepackage{amsmath}
\usepackage{amssymb}
\usepackage{booktabs}
\usepackage{multirow}
\usepackage{epstopdf}
\usepackage{authblk}

\def\argmin{\mathop{\rm argmin}}

\newcommand{\s}{\ensuremath{\mathbb{S}}}

\newcommand{\real}{\ensuremath{\mathbb{R}}}

\newcommand{\ltwo}{\ensuremath{\mathbb{L}^2}}

\newcommand{\inner}[2]{\left\langle #1,#2 \right\rangle}

\def\sl{\mathop{\mathfrak s\mathfrak l}\nolimits}

\def\argmin{\mathop{\rm argmin}}

\def \P {\mathcal{P}}

\def \T {\mathcal{T}}

\newcommand{\cC}{\ensuremath{\mathbb{C}}}

\newtheorem{thm}{Theorem}
\newtheorem{defn}{Definition}

\newtheorem{algorithm}{Algorithm}

\title{Video-Based Action Recognition Using Rate-Invariant Analysis of Covariance Trajectories}
\author[1]{Zhengwu Zhang}
\author[2]{Jingyong Su}
\author[3]{Eric Klassen}
\author[4]{Huiling Le }
\author[1]{Anuj Srivastava}
\affil[1]{Department of Statistics, Florida State University}
\affil[2]{Department of Mathematics and Statistics, Texas Tech University}
\affil[3]{Department of Mathematics, Florida State University}
\affil[4]{School of Mathematics Sciences, University of Nottingham}
\begin{document}


\maketitle

\begin{center}
\textbf{Abstract}
\end{center}
Statistical classification of actions in videos is mostly performed by 
extracting relevant features, particularly covariance features,  from image frames and studying time series associated with temporal evolutions of these features. 
A natural mathematical representation of 
activity videos is in form of parameterized trajectories on the covariance manifold, i.e. the set of 
symmetric, positive-definite matrices (SPDMs).  
The variable execution-rates of actions implies variable  parameterizations 
of the resulting trajectories, and complicates their classification. Since action classes are invariant to execution rates, 
one requires rate-invariant metrics for comparing trajectories. A recent paper represented trajectories
using their {\it transported
square-root vector fields} (TSRVFs), defined by parallel translating scaled-velocity vectors of trajectories 
to a reference tangent space on the manifold.  To avoid arbitrariness of selecting the reference and to reduce distortion 
introduced during this mapping, 
we develop a purely intrinsic approach where SPDM trajectories 
are represented by redefining their TSRVFs at the starting points of 
the trajectories, and analyzed as elements of a vector bundle on the manifold. Using a natural Riemannain metric on vector
bundles of SPDMs,
we compute geodesic paths and geodesic distances between trajectories in the quotient space of this vector bundle, with respect 
to the re-parameterization group. This makes
the resulting comparison of trajectories invariant to their re-parameterization. We demonstrate this framework on two applications involving video
classification: 
visual speech recognition or lip-reading and hand-gesture recognition. In both cases we achieve results either comparable to 
or better than the current literature. 
\vspace*{.3in}

\noindent\textsc{Keywords}: {Action recognition, covariance manifold, trajectories on manifolds, vector bundles, rate-invariant classification}

\section{Introduction}
The problem of classification of human actions or activities in video sequences is both important and challenging.
It has applications in video surveillance, lip reading, pedestrian tracking, hand-gesture recognition, manufacturing quality control, human-machine 
interfaces, and so on. Since the size of video data is generally very high, 
the task of video classification is often performed by extracting certain low-dimensional features of 
interest -- geometric, motion, colorimetric features, etc -- from each frame and then forming 
temporal sequences of these features for full videos. Consequently, analysis of videos get replaced by 
modeling and classification of longitudinal observations in a certain feature space. (Some papers discard 
temporal structure by pooling all feature together but that represents a severe loos of information.) Since many features are naturally
constrained to lie on nonlinear manifolds, the corresponding video representations form parameterized trajectories
on these manifolds. Examples of these manifolds include unit spheres, Grassmann manifolds, tensor manifolds, 
and the space of probability distributions. 

\begin{figure}
\begin{center}
\begin{tabular}{c}
\includegraphics[height=2in]{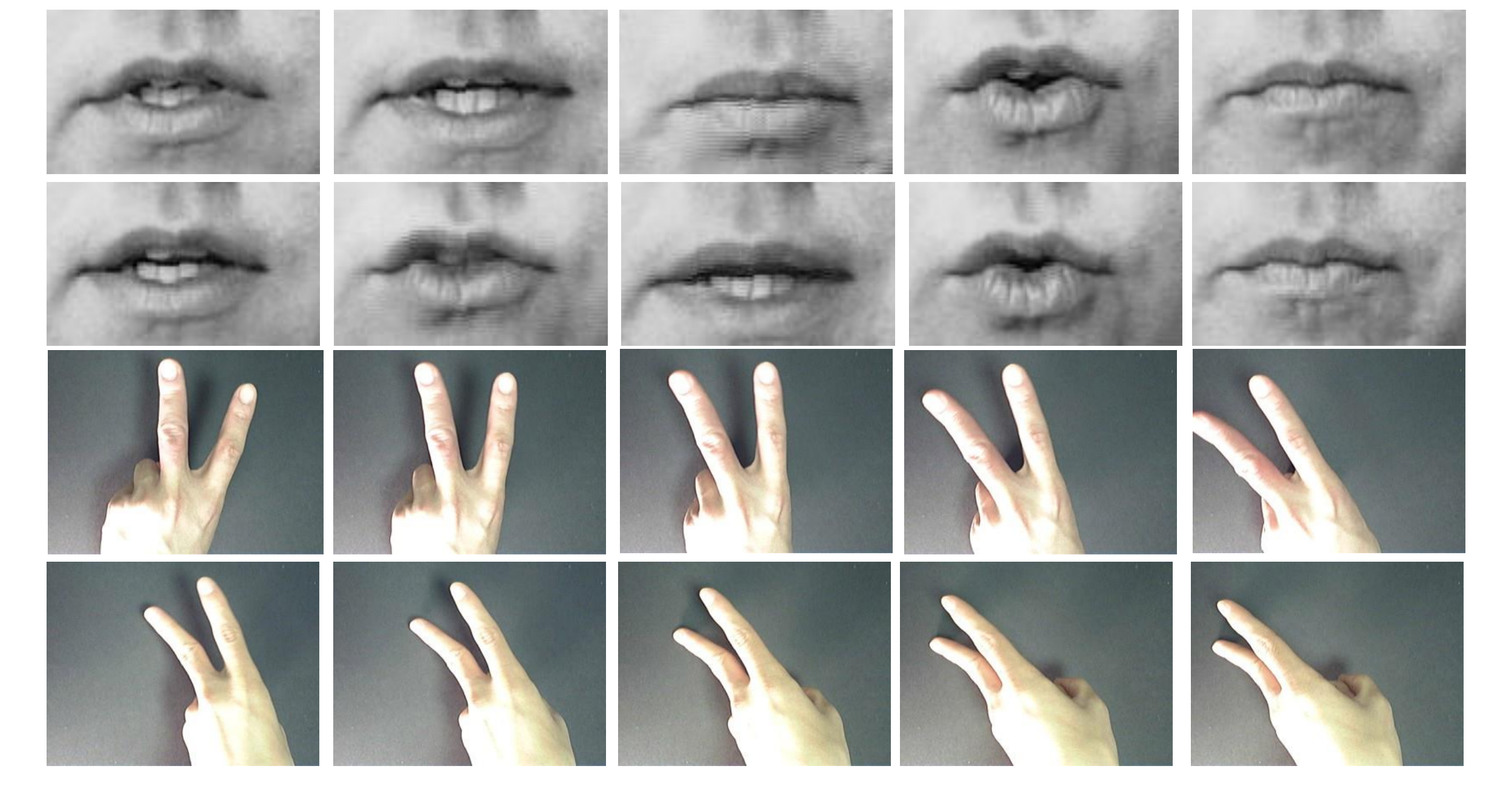}
\end{tabular}
\caption{Examples of video frames in visual-speech recognition (first two rows) and hand-gesture classification (last two).
} 
\label{fig:ill}
\end{center}
\end{figure}

One of the most commonly used and effective feature in image analysis is a covariance matrix, as shown via
applications in medical imaging \cite{Arsigny06,Pennec:2006} and computer vision \cite{Tuzel2006,Su_2014_CVPR,Harandi2012,4547427,Kim2007,Guo:2010}. 
These matrices are naturally constrained to be symmetric positive-definite matrices (SPDMs) and have also
played a prominent role as 
region descriptors in texture classification, object detection, object tracking, action recognition and face recognition.  Tuzel et al. 
\cite{Tuzel2006} introduced the concept of {\it covariance tracking}
where they extracted a covariance matrix for each video frame and studied the temporal evolution of this matrix 
in the context of pedestrian tracking in videos. Since the set of SPDMs is a nonlinear manifold, denoted by ${\tilde{\P}}$, 
a whole video segment can be represented as a (parameterized) trajectory on ${\tilde{\P}}$. 
In this paper we focus on the problem of classification of actions or activities by treating them
as parameterized trajectories on $\tilde{\P}$. The two specific applications we will study are: visual-speech recognition and 
hand-gesture classification.  Fig. \ref{fig:ill} illustrates examples of video frames for these two applications. 

One challenge in characterizing activities as trajectories comes from the 
variability in execution rates. 
The execution rate of an activity dictates the parameterization of the 
corresponding trajectory. Even for the same activity performed by the same person,  
the execution rates can potentially differ a lot.  
Different execution rate implies that the corresponding trajectories go through the same sequences of points in $\tilde{\P}$ but have different parameterizations. 
Directly analyzing such trajectories without alignment, e.g. comparing the difference, calculating point-wise mean and covariance, can be misleading (the mean is not representative of individual trajectories, and the variance is artificially inflated). 
  
To make these issues precise, we develop some notation first. Let 
$\alpha: [0,1] \to \tilde{\P}$ be a trajectory and let 
 $\gamma:[0,1] \to [0,1]$ be a positive diffeomorphism such that $\gamma(0) = 0$ and $\gamma(1) = 1$. 
This $\gamma$ 
 plays the role of a time-warping function, or a re-parameterization function, so that the composition $\alpha \circ \gamma$
 is now a time-warped or re-parameterized version of $\alpha$. In other words, the trajectory $\alpha \circ \gamma$ 
 goes through the same set of points as $\alpha$ but at a different rate (speed). 
 \begin{enumerate}
 \item {\bf Pairwise Registration}: 
Now, let $\alpha_1, \alpha_2: [0,1] \to \tilde{\P}$ be  two trajectories on $\tilde{\P}$. The process of registration of
$\alpha_1$ and $\alpha_2$ is to find a warping $\gamma$ such that $\alpha_1(t)$ is 
optimally registered to $\alpha_2(\gamma(t))$ for all $t \in [0,1]$. In order to ascribe a meaning to optimality, we
need to develop a precise criterion. 
\item {\bf Groupwise or Multiple Registration}: This problem 
can be extended to more than two trajectories: let $\alpha_1, \alpha_2,\dots, \alpha_n$ be $n$ trajectories on $\tilde{\P}$, and we
want to find out time warpings $\gamma_1, \gamma_2, \dots, \gamma_n$ such that for all $t$, the variables $\{\alpha_i(\gamma_i(t))\}_{i=1}^n$ 
are optimally registered. A solution for pairwise registration can be extended to the multiple alignment problem as 
follows -- for the given trajectories, first define a {\it template} trajectory and then align each given trajectory to this 
template in a pairwise fashion.  One way of defining this template is to use the mean of given trajectories 
under an appropriately chosen metric. 
\end{enumerate}

Notice that the problem of {\it comparisons of trajectories} is different from the problem of 
{\it curve fitting} or {\it trajectory estimation} from noisy data. 
Many papers have studied spline-type solutions
for fitting curves to discrete, noisy data points on manifolds \cite{Jupp:1987,Morris:99,Kume:2007,samir-etal-FOCM,su-etal-JIVC:2012} but in this paper we assume that
the trajectories are already available through some other means. 

\subsection{Past Work \& Their Limitations}
There are very few papers in the literature for analyzing, in the sense of comparing, averaging or clustering, trajectories
on nonlinear manifolds. 
Let $d_{\tilde{\P}}$ denote the geodesic distance resulting from the chosen Riemannian metric on $\tilde{\P}$. It can be shown that
the quantity $\int_0^1 d_{\tilde{\P}}(\alpha_1(t), \alpha_2(t)) dt$ forms a proper distance on the set $\tilde{\P}^{[0,1]}$, the space of 
all trajectories on $\tilde{\P}$. 
For example, \cite{kendall-arxiv:2014} uses this metric, combined with the arc-length distance on $\s^2$, to cluster hurricane tracks. 
However, this metric is not immune to different temporal evolutions of hurricane tracks. Handling this variability requires performing some kind
of temporal alignment. 
It is tempting to use the following modification of this distance to align two trajectories: 
\begin{equation} \label{eqn:oldalign}
\inf_{\gamma \in \Gamma} \left( \int_0^1 d_{\tilde{\P}}(\alpha_1(t), \alpha_2 (\gamma(t))) dt \right)\ ,
\end{equation}
but this can lead to degenerate solutions (also known as the {\it pinching problem}, described for 
real-valued functions in  \cite{ramsay2005}). Pinching implies that 
a severely distorted $\gamma$ is used to eliminate (or minimize) those parts of $\alpha_2$ that
do not match with $\alpha_1$, which can be done even when $\alpha_2$ is mostly different from $\alpha_1$. While this degeneracy can be avoided
using a regularization penalty on $\gamma$, some of the other problems remain, including the fact that the solution is not symmetric. 

A recent solution, presented in \cite{su2014,Su_2014_CVPR}, develops the concept of 
elastic trajectories to deal with the parameterization variability. It represents each 
trajectory by its transported square-root vector field (TSRVF) defined as: 
$$
h_{\alpha}(t) =  \left({ \dot{\alpha}(t)  \over \sqrt{ | \dot{\alpha}(t) |} } \right)_{\alpha(t) \rightarrow c} \in \T_{c}({\tilde{\P}}) \ ,
$$
where $c$ is pre-determined reference point on $\tilde{\P}$ and $\rightarrow$ denotes a parallel transport of the 
vector $\dot{\alpha}(t)$ from the point $\alpha(t)$ to $c$ {\it along a geodesic path}. This way a trajectory can be mapped
into the tangent space $\T_c(\tilde{\P})$ and one can compare/align them using the $\ltwo$ norm on that vector space. 
More precisely, the quantity $\inf_{\gamma} \| h_{\alpha_1} - h_{\alpha_2 \circ \gamma}\|$ provides not only a 
criterion for optimality of $\gamma$ but also approximates a proper metric for averaging and other statistical analyses.  
(The exact metric is based on the use of semigroup $\tilde{\Gamma}$, the set of all absolutely continuous, 
weakly increasing functions, rather than $\Gamma$. We refer the reader to  \cite{su2014} for details.) 
This TSRVF representation is an extension of the SRVF used for elastic shape analysis of curves in 
Euclidean spaces \cite{srivastava2011}.
One limitation of this framework is that the choice of reference point, $c$, is left arbitrary. The
results can potentially change with $c$ and make it difficult to interpret the results. A related, and bigger issue, is that the
transport of tangent vectors $\dot{\alpha}(t)$  to $c$, along geodesics, can introduce large distortion, especially when the trajectories
are far from $c$ on the manifold. 

\subsection{Our Approach}

We present a different approach that does not require a choice of $c$. Here the
trajectories are represented by their transported vector fields but without a need to transport them to a global reference point. 
For each trajectory $\alpha_i$, the reference
point is chosen to be its starting point $\alpha_i(0)$, and the transport is performed along the trajectory itself. In other words, for each 
$t$, the velocity vector $\dot{\alpha}_i(t)$ is transported along $\alpha$ to the tangent space of the starting point $\alpha_i(0)$. 
This idea has been used previously in \cite{Kume:2007} and others for mapping trajectories into vector spaces, and results in a 
relatively stable curve, with smaller distortion than the TSRVFs of 
\cite{su2014}. We then develop a metric-based framework 
for comparing, averaging, and modeling such curves in a manner that is invariant to their re-parameterizations. 
Consequently, this framework provides a natural solution for removal of rate, or {\it phase}, variability from trajectory data.

Another issue is the choice of Riemannian  metric on $\tilde{\P}$. 
Although a larger number of Riemannian structures and 
metrics have been used for $\tilde{\P}$ in past papers and books \cite{dryden-etal-AOAS:2009,jost2005riemannian}, they do not provide all the mathematical 
tools we will need for ensuing statistical analysis. 
Consequently,  we 
use a different Riemannian structure than those previous papers, a structure that allows relevant mathematical tools for applying the proposed framework.  

The rest of this paper paper is organized as follows. In Section \ref{sec:mf}, we introduce a framework of aligning, averaging and comparing of trajectories on a general manifold $M$. Since we mainly focus on $\tilde{\P}$, in Section \ref{sec:spd}, we introduce the details of the Riemannian structure on $\tilde{\P}$ used to implement the comparison framework described in Section \ref{sec:mf}. In Section \ref{sec:exp}, we demonstrate the proposed work with real-world action recognition data involving two 
applications: lip-reading and hand-gesture recognition. 

\section{Riemannian Structure on $\tilde{\P}$} \label{sec:spd}
Here we will discuss the geometry of  $\tilde{\P}$ and impose a Riemannian structure that facilitates 
our analysis of trajectories  on $\tilde{\P}$. Most of the background material is derived in Appendix 
\ref{sec:geom} with only the final expressions noted here. For a beginning reader in differential 
geometry, we strongly recommend reading Appendix \ref{sec:geom} first. 

We start by choosing an appropriate  {\it Riemannian metric} on $\tilde{\P}$. Then, 
from the resulting structure, we derive expressions for the following:
(1)  {\it geodesic paths} between arbitrary points on $\tilde{\P}$; (2)  {\it parallel transport} of tangent vectors along geodesic paths; 
(3) {\it exponential map}; (4) {\it inverse exponential map};  and (5) {\it Riemannian curvature tensor}. 
Several past papers have studied the space of SPDMs as a 
nonlinear manifold and imposed a metric structure on that manifold \cite{Pennec:2006,Arsigny06,dryden-etal-AOAS:2009,schwartmann-etal:AOS:2008}.  
While they mostly focus on defining distances, a few of them originate from a Riemannian structure with expressions for 
geodesics and exponential maps.  
However, they do not provide expressions for all desired items (parallel transport and Riemannian curvature tensor). 
In this section, we utilize a particular Riemannian structure on $\tilde{\P}$, which is similar but not identical to the 
one in \cite{Pennec:2006}, and is particularly convenient for our purposes. 
This Riemannian structure has been used previously for other applications such as spline-fitting in \cite{su-etal-JIVC:2012}. 

Let $\tilde{\P}$ be the space of $n \times n$ SPDMs, and let $\P$ be its subset of matrices with determinant one.
The tangent spaces of $\tilde{\P}$ and $\P$ at $I$, where $I$ is the identity matrix, are $\mathcal{T}_I( \tilde{\P}) 
 =\{ A |A^t = A \}$ and $\mathcal{T}_I(\P) \equiv  \mathfrak{p}(n) =\{ A |A^t = A \text{ and } \text{tr}(A) = 0\}$. 
The exponential map at $I$ can be shown to be
 the standard matrix exponential: 
 for any $P \in \P$ ($\tilde{P} \in \tilde{\P}$), there is $A \in \T_I(\P)$ ($\tilde{A} \in \T_I(\tilde{\P})$) such that $P = e^A$ ($\tilde{P} = e^{\tilde{A}}$), where $e^A$ denotes the matrix (Lie) exponential; this relationship is one-to-one. 
Our approach is first to identify the space ${\cal P}$ with the quotient space $SL(n)/SO(n)$ and borrow
the Riemannian structure from the latter directly. Then, we straightforwardly extend the Riemannian structure on $\cal P$ to $\tilde{{\cal P}}$.
The Riemannian geometries of $SL(n)$ and its quotient space $SL(n)/SO(n)$ are
discussed in Appendix \ref{sec:geom}. As described there, the Riemannian metric at 
any point $G$ is defined by pulling back the tangent vectors under $G^{-1}$ to $I$, and then using the $trace$ metric
(see Eqn. \ref{eqn:metric-invariance-sln}). This definition  leads to expressions for exponential map, its
inverse, parallel transport of tangent vectors, and the Riemannian curvature tensor on $SL(n)$. 
It also induces a Riemannian structure on the  quotient space $SL(n)/SO(n)$ in a natural way because it is invariant 
to the action of $SO(n)$ on $SL(n)$.

\subsection{Riemannian Structure on $\P$ }

To make the connection with $\P$, we state a useful result termed the {\it polar decomposition} of square matrices. 
Recall that for any square matrix $G \in SL(n)$, one can decompose it uniquely as $G = PS$ where $P$ is a SPDM with 
determinant one and $S \in SO(n)$.  
We note in passing that 
this fact makes $\cal P$ a section of $SL(n)$ under the action of $SO(n)$. 
(If a group acts on a manifold, then a 
section of that action is defined to be a subset of the manifold that intersects each orbit of the action in at most one point.
It is easy to see that $\cal P$ satisfies that condition for the action of $SO(n)$ on $SL(n)$. For any $SO(n)$ orbit $[G]$, $\cal P$ 
intersects that orbit in only one point given by $P$ such that $G = PS$.) We also note that
this section is {\it not} an orthogonal section since it is not perpendicular to the orbit, i.e., 
the tangent space $\T_P({\cal P})$ is not perpendicular to the tangent space 
$\T_P([G])$.
We will identify ${\cal P}$ with the quotient space $SL(n)/SO(n)$ via a map $\pi$ defined as: 
$$
\pi: SL(n)/SO(n) \to \mathcal{P},  \pi([G]) = \sqrt{{\tilde G}{ \tilde G}^t}\ ,
$$
for any $\tilde{G} \in [G]$. One can check that this map is well defined and is a diffeomorphism. 
This square-root is the symmetric, positive-definite square-root of a symmetric matrix. 
One can verify that $\pi([G])$ lies in ${\cal P}$ 
by letting $\tilde{G} = PS$ (polar decomposition), and then $\pi([G]) = \sqrt{\tilde{G}\tilde{G}^t} = \sqrt{PSS^tP} = P$. 
The inverse map of $\pi$ is given by: $\pi^{-1}(P) = [P] \equiv \{PS|S \in SO(n) \} \in SL(n)/SO(n)$. 
This establishes a one-to-one correspondence between the quotient space $SL(n)/SO(n)$ and $\P$. 
We will use the map $\pi$ to push forward the chosen Riemannian metric from the quotient space $SL(n)/SO(n)$ to $\P$. \\

\noindent {\bf Geodesic between two points on $\P$}:
With that induced Riemannian metric, we can derive the geodesic path and the geodesic 
distance between any  $P_1$ and $P_2$ in $\P$. The idea is to pullback these points 
into the quotient space using $\pi^{-1}$, compute the geodesic there and then 
map the result back to $\P$ using $\pi$. As mentioned earlier $\pi^{-1}(P_1) = [P_1]$
and $\pi^{-1}(P_2) = [P_2]$. The expression for geodesic between these points in the quotient 
space is given in Appendix \ref{sec:geom}. We compute $A_{12} \in \mathfrak{p}(n)$ such that 
$e^{A_{12}} = P_1^{-1} P_2 S_{12}$ for some $S_{12} \in SO(n)$. (Let $P_{12} = P_1^{-1} P_2 S_{12}$. Note that $P_{12} \in \P$, and the rotation matrix $S_{12}$ brings $P_1^{-1} P_2$ to $\P$.) The corresponding geodesic in 
the quotient space ($SL(n)/SO(n)$) is given by $t \mapsto [P_1 e^{tA_{12}}]$ and, therefore, the desired geodesic in 
$\P$ is 
$$
t \mapsto \pi([P_1 e^{tA_{12}}]) = \sqrt{P_1 e^{2t A_{12}} P_1}\ . \\
$$
The corresponding geodesic distance is
$
d(P_1,P_2) = d(I,P_{12}) = \|A_{12}\|. 
$ \\

\noindent {\bf Parallel transport of tangent vectors along geodesic paths on $\P$}:
To determine the parallel transport along the geodesic path $t \mapsto \pi([P_1 e^{tA_{12}}])$, we recall
from Appendix \ref{sec:geom} that the two 
orthogonal subspaces of $\mathfrak{sl}(n)$: $\mathfrak{so}(n)$ and $\mathfrak{p}(n)$, which we call 
the {\it vertical} and the {\it horizontal} tangent subspaces, respectively. We identify the tangent space $\T_P(\P)$ 
with the horizontal subspace in $\T_P(SL(n))$, so that $\T_P(\P) = \{ PA|A \in \mathfrak{p}(n)\}$. 

Now let $X \in \T_{P_1}(\P)$ be a tangent vector that needs to be translated along a geodesic path from $P_1$ to $P_2$, 
given by $ t \mapsto \pi([P_1 e^{tA_{12}}])$. 
Similar to the case of quotient space in Appendix \ref{sec:geom}, 
let $B \in \mathfrak{p}(n)$ such that $X$ is identified with ${P_1}B$. In the quotient space, the parallel transport of $X$ 
along the geodesic is a vector field $t \mapsto P_1 e^{tA_{12}} B$. In $\P$, the geodesic is obtained by the forward map $\pi$, $t \mapsto \pi([P_1 e^{tA_{12}}])$. Therefore, the parallel transport vector field in $\P$ along the geodesic is the image of $P_1 e^{tA_{12}} B$ under $d\pi$:
$$ 
t \mapsto d\pi( P_1 e^{tA_{12}} B ) = \pi([P_1 e^{tA_{12}}])  T(t) B T(t)^t,
$$
where $T(t) \in SO(n)$ and $\pi([P_1 e^{tA_{12}}]) T(t) = P_1 e^{tA_{12}}$.\\

\noindent {\bf Riemannian curvature tensor on $\P$}: 
Let $X,Y$ and $Z$ be three vectors on $\T_P(\P)$, then
the Riemannian curvature tensor on $\P$ with these arguments can be calculated in the follows. Let $A,B$ and $C$ be elements in $\mathfrak{p}(n)$ such that $X = PA$, $Y = PB$ and $Z = PC$, the tensor is given by: 
$$R(X,Y)(Z) = -[[X,Y],Z] = -P[[A,B],C].$$

\noindent
We summarize these mathematical tools for trajectories analysis on the manifold $\P$:
\begin{enumerate}
\item {\bf Exponential map:} Give a point $P \in \mathcal{P}$ and a tangent vector $V \in \mathcal{T}_P(\P)$, the exponential map is given as:
$\exp_P(V) = \sqrt{Pe^{2P^{-1}V}P}.$

\item {\bf Geodesic distance}: For any two points $P_1, P_2 \in \P$, the geodesic distance between them in is given by: 
$d_{\P}(P_1, P_2)= d_{\P}(I,P_{12}) = \|A_{12}\|$ where $e^{A_{12}} = P_{12} \in \P$ and $ P_{12} = \sqrt{P_1^{-1} P_2^2 P_1^{-1} } $. 

\item {\bf Inverse exponential map:} For any $P_1,P_2 \in \P$, the inverse exponential map can be calculated using the formula: $\exp^{-1}_{P_1}(P_2) = P_1 \log(\sqrt{P_1^{-1} P_2^2 P_1^{-1}}).$

\item {\bf Parallel transport:} Given $P_1,P_2 \in \P$ and a tangent vector $V \in \mathcal{T}_{P_1}(\P)$, the tangent vector at $P_2$ which is the parallel transport of $V$ along the shortest geodesic from $P_1$ to $P_2$ is: $ P_2T_{12}^TBT_{12}$, where $B = P_1^{-1}V,T_{12}=P_{12}^{-1}P_1^{-1}P_2$ and $P_{12} = \sqrt{P_1^{-1} P_2^2 P_1^{-1} }$.

\item {\bf Riemannian curvature tensor}: For any point $P \in \mathcal{P}$ and tangent vectors $X,Y$ and $Z \in \mathcal{T}_P(\P)$, the Riemannian curvature tensor $R(X,Y)(Z)$ is given as: $R(X,Y)(Z) = -P[[A,B],C]$, where $ A = P^{-1}X , B = P^{-1}Y, C = P^{-1}Z$ and $[A,B] = AB - BA$.
\end{enumerate} 

\subsection{Extension of Riemannian Structure to $\tilde{\P}$ }
We now extend the Riemannian structure on $\P$ to $\tilde{\P}$. 
Since for any $\tilde{P} \in \tilde{\P}$ we have $\det({\tilde{P}})>0$, we can express $\tilde{P} = (P, {1 \over n} \log(\det(\tilde{P})))$ with 
$P = {\tilde{P} \over \det(\tilde{P})^{1/n}} \in \P$. Thus,  $\tilde{\P}$ is identified with the product space of $\P \times \real$. 
Moreover, for any smooth function $\psi$ on $\P$, the Riemannian metric $\tilde{g}_{(P,x)} = \psi^2dx^2 + g_P$, where $\tilde{g}_{(P,x)}$ and $g_P$ denote the Riemannian metrics on $\tilde{\P}$ and $\P$ respectively, gives $\tilde{\P}$ the structure of a ``warped'' Riemannian product. In this paper we set $\psi \equiv \sqrt{n}$.  \\

\noindent {\bf Geodesic between points on $\tilde{\P}$}:
Using the above Riemannian metric ($\tilde{g}_{(P,x)}$) on $\tilde{\P}$, the resulting squared distance on $\tilde{P}$ between $I$ and $\tilde{P}$ is: 
$$d_{\tilde{\P}}(I,\tilde{P})^2 = d_{\P}(I,P)^2 + \frac{1}{n}(\log(\det(\tilde{P})))^2.$$
For two points $\tilde{P}_1$ and $\tilde{P}_2$, let $\tilde{P}_{12} = \tilde{P}_1^{-1} \tilde{P}_2S_{12}$ for some $S_{12} \in SO(n)$, and $\tilde{P}_{12} \in \tilde{\P}$, we have $\det(\tilde{P}_{12}) = \det(\tilde{P}_2)/\det(\tilde{P}_1)$. Therefore, the resulting squared geodesic distance between two SPDMs $\tilde{P}_1$ and $\tilde{P}_2$ is
$d_{\tilde{\P}}(\tilde{P}_1,\tilde{P}_2)^2 = d_{\tilde{\P}}(I,\tilde{P}_{12})^2$
$$ = d_{\P}(I,P_{12})^2 +\frac{1}{n}(\log(\det(\tilde{P}_2))-\log(\det(\tilde{P}_1)))^2.$$ 

Next, let $\phi = (\phi_1, \phi_2)$ denote the geodesic in $\tilde{\P}$, where $\phi_1$ is a geodesic on $\P$ and $\phi_2$ is a geodesic in $\real$. The geodesic $\phi_1$ from $P_1$ to $P_2$ in $\P$ is given by 
$\phi_1(t) = \sqrt{P_1 e^{2t A_{12}} P_1}$ from earlier derivation, and the geodesic $\phi_2$, a line segment from $\frac{1}{n}\log(\det(\tilde{P}_1))$ to $\frac{1}{n}\log(\det(\tilde{P}_2))$, is 
$\phi_2(t) = (1-t)\frac{1}{n}\log(\det(\tilde{P}_1)) + t  \frac{1}{n}\log(\det(\tilde{P}_2))$. Therefore, with simple calculation, the matrices in $\tilde{\P}$ corresponding to the geodesic path $\phi$ are $\phi(t) = \det(\tilde{P}_1)^{1/n}\left( \frac{\det(\tilde{P}_2)}{\det(\tilde{P}_1)} \right)^{t/n} \sqrt{P_1 e^{2t A_{12}} P_1}  \in \tilde{\P}$. \\

\noindent {\bf Parallel transport of tangent vectors along geodesic paths on $\tilde{\P}$}:
If $\tilde{V}$ is a tangent vector to $\tilde{\P}$ at $\tilde{P}$, where $\tilde{P}$ is identified with $\tilde{P} = (P,x)$ and $x = \frac{1}{n} \log(\det(\tilde{P}))$, we can express $\tilde{V}$ as $\tilde{V} = (V,v)$ with $V$ being a tangent vector of $\P$ at $P$. The parallel transport of $\tilde{V}$ is the parallel transport of each of its two components in the corresponding space, and the transport of $v$ is itself. The other part, i.e. the parallel translation of $V$ in $\P$ has been dealt earlier. \\

\noindent {\bf Riemannian curvature tensor on $\tilde{\P}$}:
By identifying the tangent vector $\tilde{V} = (V,v)$, the non-zero curvature tensors are just the part in $\P$. Therefore, the curvature tensor for $\tilde{X} = (X,x)$, $\tilde{Y}=(Y,y)$ and $\tilde{Z}=(Z,z)$ on $\T_{\tilde{P}}(\tilde{\P})$ is the same as $R(X,Y)Z$. 

\noindent
Again, we summarize these mathematical tools on $\tilde{\P}$:
\begin{enumerate}
\item {\bf Exponential map:} Give $\tilde{P} \in \tilde{\P}$ and a tangent vector $\tilde{V} \in \T_{\tilde{P}}(\tilde{\P})$. We denote $\tilde{V} = (V,v)$, where $V \in \T_P(\P)$, $P = \tilde{P}/\det(\tilde{P})^{1/n}$ and $v = \frac{1}{n}\log(\det(\tilde{P}))$. The exponential map $\exp_{\tilde{P}}(\tilde{V})$ is given as  $e^v\exp_P(V)$, where $\exp_P(V) = \sqrt{Pe^{2P^{-1}V}P}.$

\item {\bf Geodesic distance}: For any $\tilde{P}_1, \tilde{P}_2 \in \tilde{\P}$, the squared geodesic distance between them is : 
$d_{\tilde{\P}}(\tilde{P}_1, \tilde{P}_2)^2= d_{\P}(I,P_{12})^2 + \frac{1}{n}(\log(\det(\tilde{P}_2))-\log(\det(\tilde{P}_1)))^2$, where $ P_{12} = \sqrt{P_1^{-1} P_2^2 P_1^{-1} } $. 

\item {\bf Inverse exponential map:} For any $\tilde{P}_1,\tilde{P}_2 \in \tilde{\P}$, the inverse exponential map $\exp^{-1}_{\tilde{P}_1}(\tilde{P}_2) = \tilde{V} \equiv (V, v)$, where $V = P_1 \log(\sqrt{P_1^{-1} P_2^2 P_1^{-1}})$ and $v = \frac{1}{n}\log(\det(P_2)) - \frac{1}{n}\log(\det(P_1)) $. 

\item {\bf Parallel transport:} For any $\tilde{P}_1,\tilde{P}_2 \in \tilde{\P}$ and a tangent vector $\tilde{V} = (V,v) \in \mathcal{T}_{\tilde{P}_1}(\tilde{\P})$, the parallel transport of $\tilde{V}$ along the geodesic from $\tilde{P}_1$ to $\tilde{P}_2$ is: $ (P_2T_{12}^TBT_{12},v)$, where $B = P_1^{-1}V,T_{12}=P_{12}^{-1}P_1^{-1}P_2$ and $P_{12} = \sqrt{P_1^{-1} P_2^2 P_1^{-1} }$.

\item {\bf Riemannian curvature tensor}: For tangent vectors $\tilde{X}=(X,x),\tilde{Y} = (Y,y)$ and $\tilde{Z} = (Z,z) \in \mathcal{T}_{\tilde{P}}(\tilde{\P})$, the Riemannian curvature tensor $R(\tilde{X},\tilde{Y})(\tilde{Z})$ is the same as $R(X,Y)(Z)$.
\end{enumerate}

\subsection{Differences From Past Riemannian Structures}
Pennec et al.  \cite{Pennec:2006} and others \cite{Fstner99,Fletcher04principalgeodesic,Tuzel:2008} and even 
others before them have utilized
a Riemannian structure on $\P$ that is commonly known as {\it the 
Riemanian metric} in the literature.
Here we clarify the difference between our Riemannian structure and the structure used in these past papers. 

We have used a natural bijection $\pi:SL(n)/SO(n) \to \P$ given by $\pi([G])= \sqrt{GG^{T}}$ to inherit the 
Riemannian structure from the quotient space to $\P$. Its inverse is given by $\pi^{-1}(P) = [P]$. 
There is another bijection between these same two sets, defined by $\pi_B([G]) = GG^T$, and with 
the inverse $\pi_B^{-1}(P) = [\sqrt{P}]$. (Note that an element of ${\cal P}$ can be raised to any positive power in a well-defined way by diagonalizing.)
If we use $\pi_B$ to transfer the Riemannian structure from 
the quotient space $SL(n)/SO(n)$ onto $\P$, we will reach the one used in earlier papers. In other words, 
the structure on the quotient space is the same for both cases, only the inherited structure on $\P$ is 
different due to the different bijections. 
 Note that for all $P\in{\cal P}$, $\pi_B(\pi^{-1}(P))=P^2$. It follows that the map from ${\cal P} \to{\cal P}$ defined by $P\mapsto P^2$ 
 forms an isometry with our metric on one side and the older metric on the other.

The Riemannian metric on $SL(n)$ (trace metric on pulled-back matrices at
identity) is invariant under right translation by elements of $SO(n)$ 
and under left translation by elements of $SL(n)$. It follows that the induced metric on the quotient space 
is invariant under the natural left action of $SL(n)$ on $SL(n)/SO(n)$. (This action  is given by $(g,[G])\mapsto [gG])$.) 
It is a well-known fact that, up to a fixed scalar multiple, this is the only metric on $SL(n)/SO(n)$ that is invariant under the action by $SL(n)$. 
This aspect is same for both the cases. 
If we push forward the action of $SL(n)$ on the quotient space via $\pi$, the resulting action of $SL(n)$ on ${\cal P}$ is given by $(G,P)\mapsto \sqrt{GP^2G^T}$,
but if we push forward the action via $\pi_B$, the resulting action of $SL(n)$ on ${\cal P}$ is given by $(G,P)\mapsto GPG^T$. 

Given a simple relationship between the two Riemannian structures, one may conjecture
that the relevant formulae (parallel transport, curvature tensor, etc) can be mapped from 
one setup to another. While this is true in principle, the practice is much harder since 
it involves the expression for the differential 
of this map which is somewhat complicated. 

\section{Analysis of Trajectories on $\tilde{\P}$} \label{sec:mf}
Now that we have expressions for calculating relevant geometric quantities on $\tilde{\P}$, we
return to the problem of analyzing trajectories on $\tilde{\P}$. In the following we derive the framework for a
 general Riemannian manifold $M$, keeping in mind that $M = \tilde{\P}$ in our applications. 
 
\subsection{Representation of Trajectories }
Let $\alpha$ denote a smooth trajectory on a Riemannian manifold $M$ and ${\cal F}$ denote the set of all such trajectories: ${\cal F} =
\{ \alpha:[0,1] \to M | \alpha\ \ \mbox{is smooth}\}$. Define
$\Gamma$ to be the set of all orientation preserving diffeomorphisms
of $[0,1]$: $\Gamma=\{\gamma:[0,1]\to [0,1]|\gamma(0)=0,\
\gamma(1)=1,\ \gamma\,$ is a diffeomorphism$\}$. $\Gamma$ forms a group under the composition operation. If
$\alpha$ is a trajectory on $M$, then $\alpha \circ \gamma$ is a
trajectory that follows the same sequence of points as $\alpha$ but
at the evolution rate governed by $\gamma$. More technically, the
group $\Gamma$ acts on ${\cal F}$, ${\cal F} \times \Gamma \to {\cal
F}$, according to  $(\alpha,\gamma)=\alpha\circ\gamma$.

We introduce a new representation of trajectories that will be used to compare and register them. 
We assume that for any two points $\alpha(\tau_1),\alpha(\tau_2)  \in M, \tau_1 \neq \tau_2$, 
we have a mechanism for parallel transporting any vector
$v \in \T_{\alpha(\tau_1)}(M)$ {\bf along} $\alpha$ from $\alpha(\tau_1)$ to $\alpha(\tau_2)$, denoted by $(v)_{\alpha(\tau_1) \rightarrow \alpha(\tau_2)}$.

\begin{defn} \label{defn}
Let $\alpha: [0, 1] \rightarrow M$ denote a smooth trajectory starting with $p=\alpha(0)$.
Given a trajectory $\alpha$, and the velocity vector field $\dot{\alpha}$,  define its transported square-root vector field (TSRVF) to
be a scaled parallel-transport of the vector field along $\alpha$ to the starting point $p$ according to: for each $\tau \in [0,1]$, 
$q(\tau) = ({ \dot{\alpha}(\tau) \over \sqrt{ | \dot{\alpha}(\tau) |} })_{\alpha(\tau) \rightarrow p}  \in \T_{p}(M) \ $, where
$| \cdot |$ denotes the norm that is defined through the Riemannian metric on $M$. 
\end{defn}
Note  the difference in this definition from the one in \cite{su2014} where the parallel transport was along geodesics to a reference point $c$. Here
the parallel transport is along $\alpha$ and to the starting point $p$. The concept of parallel transportation 
of velocity vector along trajectories has also been used previously in \cite{Kume:2007} and others. This reduces distortion in 
representation relative to the parallel transport of  \cite{su2014} to a faraway reference point. 

This TSRVF representation maps a trajectory $\alpha$ on 
$M$ to a curve $q$ in $\T_p(M)$. What is the range space of all such mapping? 
For any point $p \in M$,  denote the set of all square-integrable curves in $\T_p(M)$ as $\cC_p \equiv \ltwo([0, 1], \T_p(M))$. 
The space of interest, then,  becomes an infinite-dimensional vector bundle 
$\cC = \coprod_{p \in M} \cC_p$, which is the 
indexed union of $\cC_p$ for every $p\in M$. 
Note that the TSRVF representation is bijective:  any trajectory $\alpha$ is uniquely represented by a pair $(p,q(\cdot)) \in \cC$, 
where $p \in M$ is the starting point, $q \in \cC_p$ is its TSRVF.  We can reconstruct the trajectory from $(p,q)$ using the covariant integral defined
later (in Algorithm \ref{algo1}).

\subsection{Riemannian Structure on $\cC$}
In order to compare trajectories, we will compare their corresponding representations in $\cC$ and that 
requires a Riemannian structure on the latter space. 
Let $\alpha_1, \alpha_2$  be two trajectories on $M$, with starting points $p_1$ and $p_2$, respectively, 
and let the corresponding TSRVFs be $q_1$ and $q_2$. Now $\alpha_1, 
\alpha_2$ are represented as two points in the vector bundle $(p_1,q_1),(p_2,q_2) \in \cC$ over $M$. 
This representation space is an infinite-dimensional bundle, whose fiber over each point $p$ in $M$ is $\cC_p$. 

We impose the following Riemannian structure on $\cC$. 
For an element $(x,v(\cdot))$ in $\cC$, where $x\in M$, $v \in \cC_x$, we naturally identify the tangent space at $(x,v)$ to be: $\T_{(x,v)}{(\cC)} \cong \T_x(M)\oplus \cC_x$. To see this, suppose we have a curve in $\cC$ given by $(x(s),v(s,\tau))$, $s, \tau \in [0,1]$. The velocity vector to this curve at $s=0$ is given by $(x_s(0),\nabla_{x_s}v(0,\cdot))\in \T_x(M)\oplus 
\cC_x$, where $x_s$ denotes $dx/ds$, and $\nabla_{x_s}$ denotes \emph{ covariant differentiation of tangent vectors}. The Riemannian inner product on $\cC$ is defined in an obvious way: If $(u_1,w_1(\cdot))$ and $(u_2,w_2(\cdot))$ are both elements of $\T_{(x,v)}(\cC)\cong
 \T_x(M)\oplus \cC_x$, define 
\begin{equation} \label{eqn:ccrp}
\langle (u_1,w_1(\cdot)),(u_2,w_2(\cdot))\rangle= (u_1\cdot u_2) +\int_0^1 (w_1(\tau)\cdot w_2(\tau))\,d\tau ,
\end{equation}
where the inner products on the right denote the original Riemannian metric in $\T_x(M)$.

Next, for given two points $(p_1,q_1)$ and $(p_2,q_2)$ on $\cC$, we are interested in finding the geodesic path 
connecting them. Let $\left( x(s),v(s) \right), s\in[0,1]$ be a path with 
$\left( x(0),s(0) \right) = (p_1, q_1)$ and $\left( x(1),s(1) \right) = (p_2, q_2)$.  We have the following 
characterization of geodesics on $\cC$.

\begin{thm} \label{thm:geodesicequ}
A parameterized path  $[0,1]\to \cC$ given by $s \mapsto (x(s),v(s,\tau))$ on $\cC$ (where the variable $\tau$ corresponds to the parametrization in $\cC_x$),  
is a geodesic on $\cC$ if and only if:
\begin{equation} \label{gdequ1}
\begin{array}{lcl}
\nabla_{x_s}x_s + \int _0^1 R(v,\nabla_{x_s} v)(x_s)d\tau&=&0  \quad \text{  for every } s,\\
\nabla_{x_s}(\nabla_{x_s}v)(s,\tau)&=&0 \quad \text{  for every } s, \tau.
\end{array}
\end{equation}
Here $R(\cdot,\cdot)(\cdot)$ denotes the Riemannian curvature tensor, 
$x_s$ denotes $dx/ds$, and $\nabla_{x_s}$ denotes the covariant differentiation of tangent vectors on tangent space $\T_{x(s)}(M)$.
\end{thm}
\noindent
{\bf Proof}: We will prove this theorem in two steps. \\
(1) First, we consider with a simpler case where the space of interest is the tangent bundle $TM$ of the Riemannian manifold $M$. 
An element of $TM$ is denoted by $(x,v)$, where $x\in M$ and $v \in \T_x(M)$.  It is natural to identify $\T_{(x,v)}(TM)\cong
 \T_x(M) \oplus \T_x(M)$. 
 The Riemannian inner product on $TM$ is defined in the obvious way: If $(u_1,w_1)$ and $(u_2,w_2)$ are both elements of $\T_{(x,v)}(TM)$, define 
 $$\langle(u_1,w_1),(u_2,w_2)\rangle=u_1\cdot u_2+w_1\cdot w_2
 $$
 and, again, the inner products on the right denote the original Riemannian metric on $\T_x(M)$.
 Suppose we have a path in $I\to TM$ given by $s\mapsto (x(s),v(s))$. We define the energy of this path by 
 $$E=\int_0^1 (x_s\cdot x_s +  \nabla_{x_s}v\cdot \nabla_{x_s}v) ds.$$
 The integrand is the inner product of the velocity vector of the path with itself. It is a standard result that a geodesic on $TM$ can be characterized as a path that is a critical point of this energy function on the set of all paths between two fixed points in $TM$. To derive local equations for this geodesic, we now assume we have a parameterized family of paths denoted by $(x(s,t),v(s,t))$, where $s$ is the parameter of each individual path in the family (as above) and the variable $t$ tells us which path in the family we are in. Assume $0\leq s\leq 1$ and $t$ takes values on $(-\delta,\delta)$ for some small $\delta$. We want all the paths in this family to start and end at the same points of $TM$, so assume that $(x(0,t),v(0,t))$ and $(x(1,t),v(1,t))$ are constant functions of $t$. The energy of the path with index $t$ is given by:  
 $$
 E(t)=\int_0^1 (x_s\cdot x_s +  \nabla_{x_s}v\cdot \nabla_{x_s}v) ds\ .
 $$
 To simplify notation in what follows, we will write $\nabla_s$ for $\nabla_{x_s}$ and $\nabla_t$ for $\nabla_{x_t}$. To establish conditions for $(x,v)$ to be critical, we take the derivative of $E(t)$ with respect to $t$ at $t=0$:
 $$
 E'(0)=2\int_0^1 [(\nabla_tx_s\cdot x_s)+(\nabla_t(\nabla_s v)\cdot\nabla_s v)]ds\ .
 $$
We will use two elementary facts: (a) $\nabla_t(x_s)=\nabla_s(x_t)$ and 
(b)  $R(x_t,x_s)(v)=\nabla_t(\nabla_s v)-\nabla_s(\nabla_t v)$,  without presenting their proofs. 
%
Plugging these facts into the above calculation then makes $E'(0)$ equal to: 
\setlength{\arraycolsep}{0.0em}
\begin{eqnarray*}
& &2\int_0^1[\nabla_s x_t\cdot x_s+R(x_t,x_s)(v)\cdot\nabla_sv+\nabla_s(\nabla_tv)\cdot\nabla_sv]ds\\
  & = &       2\int_0^1 [(-\nabla_sx_s\cdot x_t)+R(x_t,x_s)(v)\cdot\nabla_sv+(-\nabla_s(\nabla_sv)\cdot\nabla_t v)]ds.
\end{eqnarray*}
The second equality comes from using integration by parts on the first and third term, taking into account the fact that $x_t$ and $\nabla_tv$ vanish at $s=0,1$, 
(since all the paths begin and end at the same point). Now, using the standard identities 
$R(X,Y)(Z)\cdot W= R(Z,W)(X)\cdot Y$ and $R(X,Y)(Z)\cdot W=-R(X,Y)(W)\cdot Z$,  
we obtain: $E'(0)$ equals
\begin{eqnarray*}
&&2\int_0^1 [(-\nabla_sx_s\cdot x_t)+(-R(v,\nabla_sv)(x_s)\cdot x_t) \\
&& \hspace*{0.5in}+(-\nabla_s(\nabla_sv)\cdot\nabla_t v)]ds \\
         &=& -2\int_0^1[(\nabla_sx_s+R(v,\nabla_sv)(x_s))\cdot x_t+(\nabla_s(\nabla_sv)\cdot \nabla_tv)]ds \\
          &=& -2\int_0^1 (\nabla_sx_s+R(v,\nabla_sv)(x_s))\cdot x_t \,ds -2\int_0^1 \nabla_s(\nabla_sv)\cdot \nabla_tv\, ds\ .
\end{eqnarray*}

Now, $(x(s),v(s))$ is critical for $E$ if and only if $E'(0)=0$ for every possible variation $x_t$ of $x$ and $\nabla_t(v)$ of $v$, which is clearly true if and only if 
$$\nabla_sx_s+R(v,\nabla_sv)(x_s)=0\hbox{ and }\nabla_s(\nabla_sv)=0.$$
Thus we have derived the geodesic equations for $TM$.

\noindent
(2) Now we consider the case of the infinite dimensional vector bundle $\cC \to M$ whose fiber over $x \in M$ is $\ltwo(I,\T_x(M))$, $I = [0,1]$. A point in $\cC$ is denoted by $(x,v(\tau))$, where the variable $\tau$ corresponds to the $I$-parameter in $\ltwo(I,\T_x(M))$. The tangent space to $\cC$ at $(x,v(\tau))$ is $\T_x(M)\oplus \ltwo(I,\T_x(M))$. Suppose $(u_1,w_1(\tau))$ and $(u_2,w_2(\tau))$ are elements of this tangent space and we use the Riemannian metric:
$$\langle (u_1,w_1(\tau)),(u_2,w_2(\tau))\rangle=u_1\cdot u_2+\int_0^1 w_1(\tau)\cdot w_2(\tau)\,d\tau.$$

Now we want to work out the local equations for geodesics in $\cC$. A path in $\cC$ is denoted by $(x(s),v(s,\tau))$. The energy calculation is basically the same as above but surround everything with integration with respect to $\tau$. So, it starts out with
\begin{eqnarray*}
E&=&\int_0^1 \left(x_s\cdot x_s + \int_0^1 \nabla_{s}v\cdot \nabla_{s}v \,d\tau\right) ds\\
&=&\int_0^1\int_0^1\left(x_s\cdot x_s+\nabla_{s}v\cdot \nabla_{s}v\right)\,dsd\tau.
\end{eqnarray*}
(Of course $x_s\cdot x_s$ does not involve the parameter $\tau$, but surrounding it with $\int_0^1\dots d\tau$ does not change its value!)

In order to do the variational calculation, we now consider a parametrized family of such paths, denoted by $(x(s,t),v(s,t,\tau))$ where we assume that $x(0,t)$ and $x(1,t)$ are constant functions of $t$, and for each $\tau$, $v(0,t,\tau)$ and $v(1,t,\tau)$ are constant functions of $t$, since we want every path in our family to start and end at the same points of $\cC$.

Then, following through the computation exactly as in earlier case, we obtain
\begin{eqnarray*}
E'(0)&=&-2\int_0^1 \left(\nabla_sx_s+\int_0^1R(v,\nabla_sv)(x_s)\,d\tau\right)\cdot x_t \,ds \\
&& -2\int_0^1\int_0^1 \nabla_s(\nabla_sv)\cdot \nabla_tv\, d\tau ds.
\end{eqnarray*}
In order for our path $(x(s),v(s,\tau))$ to be critical for $E$, $E'(0)$ must vanish for every variation $x_t(s)$ of $x(s)$ and $\nabla_t(v(s,\tau))$ of $v(s,\tau)$, which is clearly true if and only if 
\begin{eqnarray*}
&&\nabla_sx_s+\int_0^1R(v,\nabla_sv)(x_s)\,d\tau = 0,\ \hbox{ for every }s \\
&&\nabla_s(\nabla_sv) = 0,\ \hbox{ for every }s\hbox{ and every }\tau\ .
\end{eqnarray*}
\begin{flushright}
{\bf Q.E.D} 
\end{flushright}

\begin{figure}[]
\begin{center}
\begin{tabular}{|ccc|}
\hline
\includegraphics[height=1.5in]{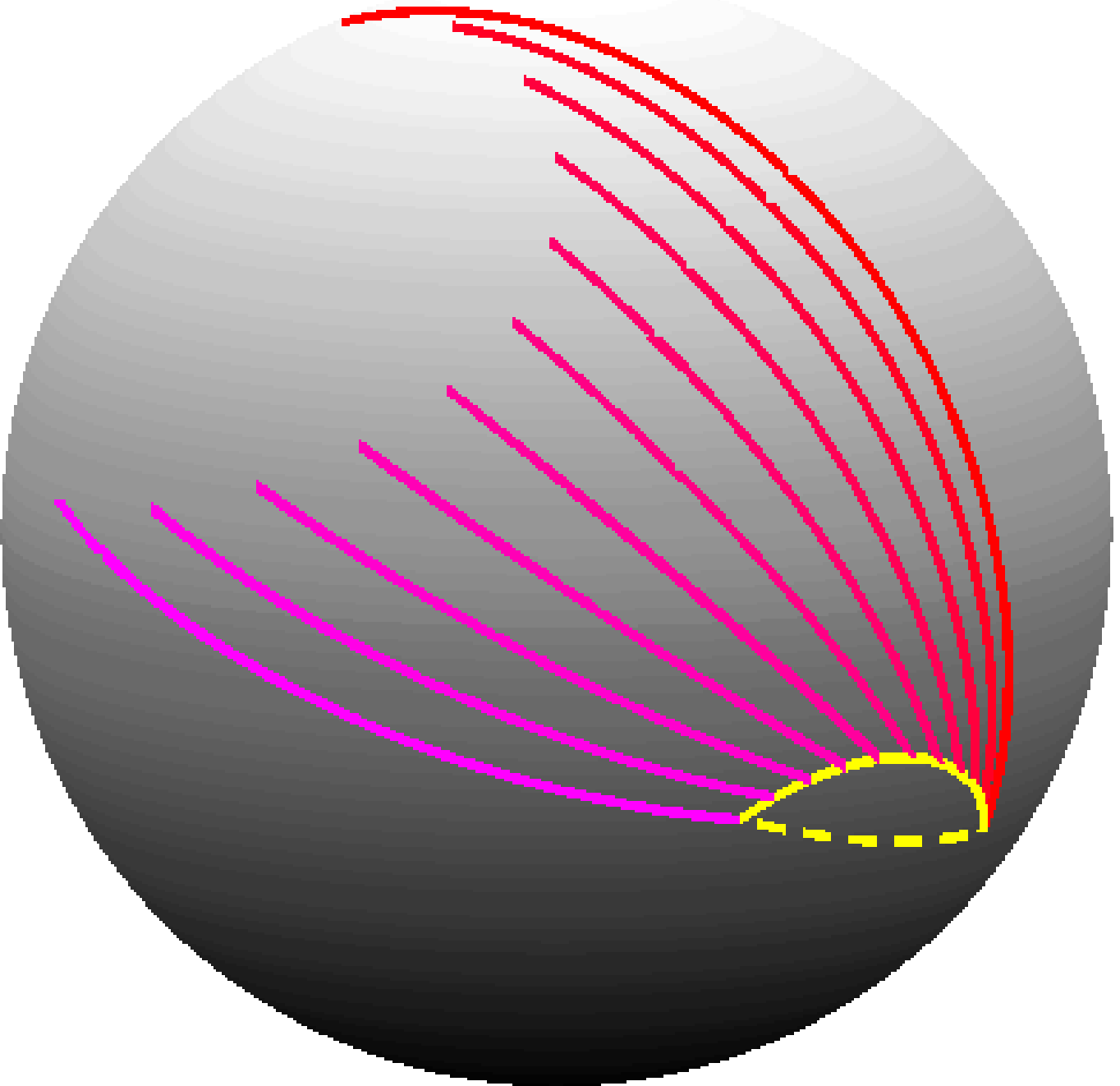}&
\includegraphics[height=1.5in]{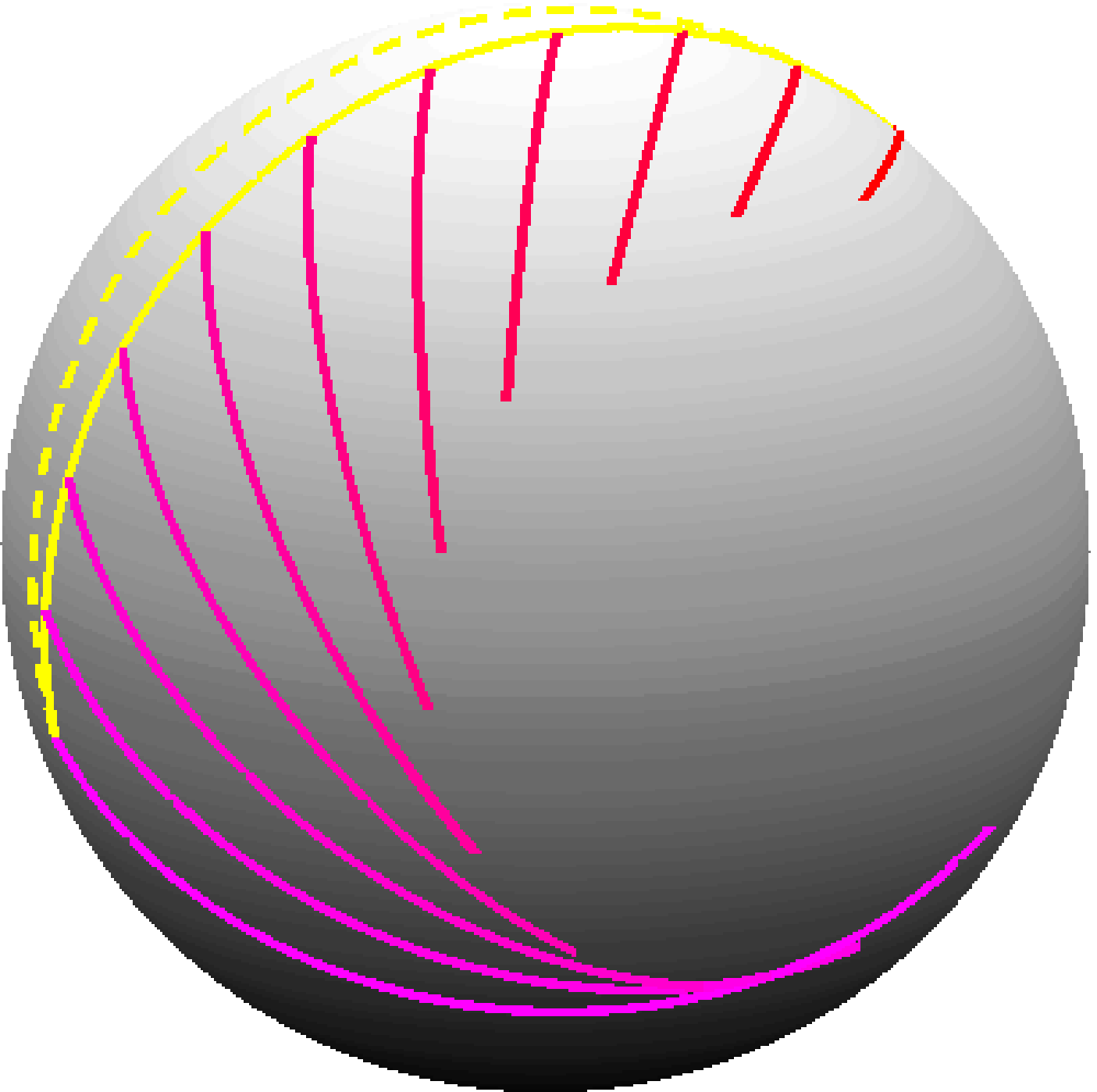}&
\includegraphics[height=1.5in]{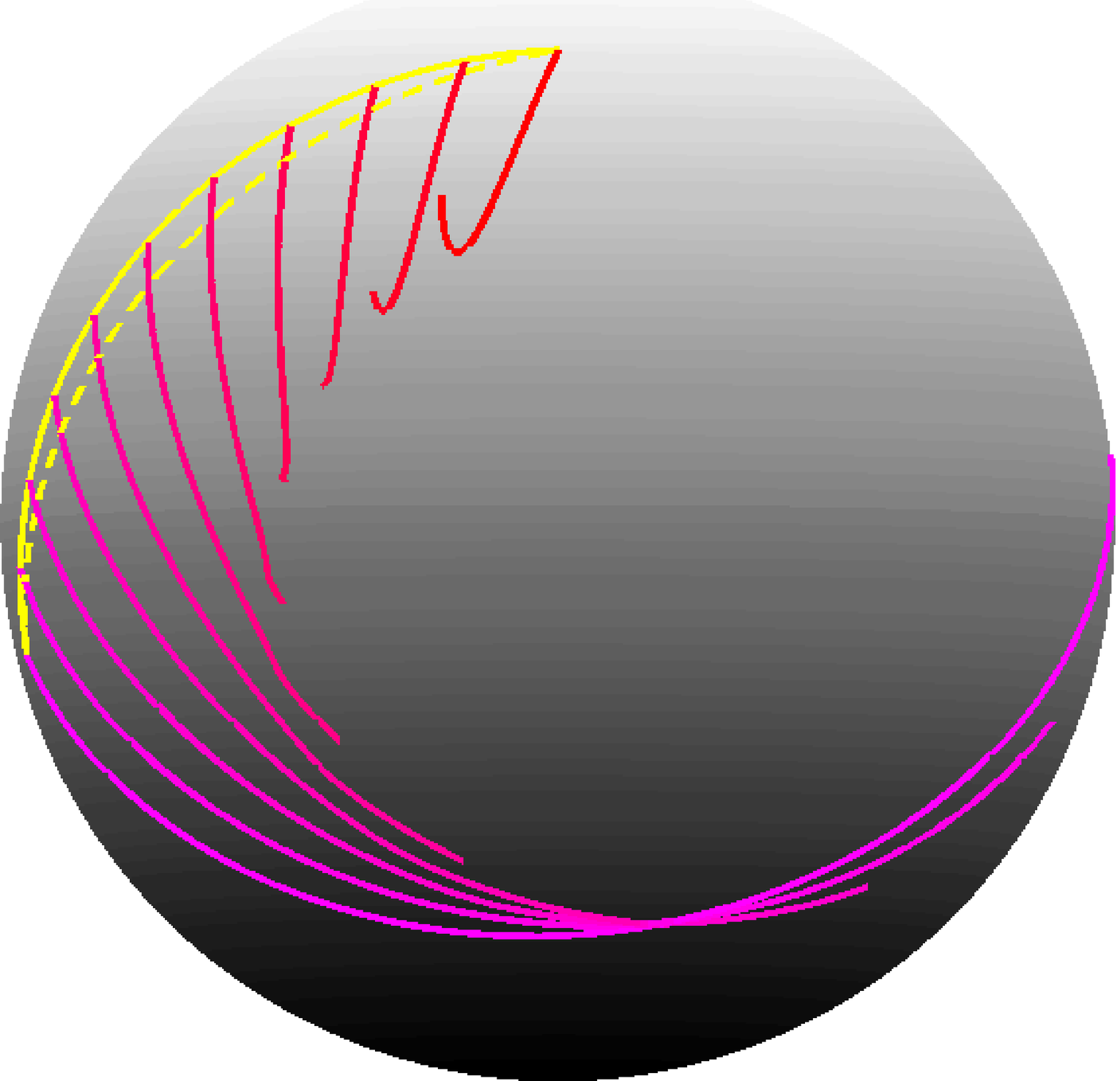}\\
\hline
\end{tabular}
\caption{Examples of geodesic between two trajectories on $\s^2$.  
}\label{fig:spheresim} 
\end{center}
\end{figure}

The geodesic path 
$(x(s),v(s))$ can be intuitively understood as follows: (1) $x(s)$ is a baseline curve on $M$ connecting $p_1$ and $p_2$, and the covariant differentiation of $x_s$ at the tangent space of $\T_{x(s)}(M)$ equals the negative integral of the Riemannian curvature tensor $ R(v(s,\tau),\nabla_{x_s} v(s,\tau))(x_s)$ with respect to $\tau$ . In other words,  values of $v$ at each $\tau$ equally determine the geodesic acceleration of $x(s)$ in the first equation. (2) The second equation leads to a fact that $v$ is covariant linear, i.e. $v(s,\tau) = a(s,\tau)+sb(s,\tau)$ and $\nabla_{x_s} a = \nabla_{x_s} b = 0$ for every $s$ and $\tau$. For a geodesic path connecting $(p_1,q_1)$ and $(p_2,q_2)$, it is natural 
to let $a(s,\tau) = q_1(\tau)_{x(0) \rightarrow x(s)}$ and $b(s,\tau) = w(\tau)_{x(0) \rightarrow x(s)}$, where $q_1(\tau)_{x(0) \rightarrow x(s)}$ and $w(\tau)_{x(0) \rightarrow x(s)}$ represent the parallel transport of $q_1(\tau)$ and $w(\tau)$ along $x(0)$ to $x(s)$, and $w$ is the difference between the TSRVFs $q_2$ and $q_1$ in $\T_{x(0)}(M)$, 
defined as  $(q_2)_{x(1) \rightarrow x(0)} - q_1$. In Fig. \ref{fig:spheresim}, we illustrate the geodeisc path between two trajectories on a simpler manifold $M = \s^2$. In each case, the yellow solid line denotes the baseline $x(s)$ and the intermediate lines are the covariant integrals (in Algorithm \ref{algo1}) of $v(s)$ with starting point $x(s)$. As comparison, the dash yellow line shows the geodesic between the starting points $p_1$ and $p_2$ on $\s^2$.

Theorem \ref{thm:geodesicequ} is only a characterization of geodesics but does not
provide explicit expressions for them. In the following section, we develop a numerical solution for constructing 
geodesics on $\cC$.

\subsection{Numerical Computation of Geodesic in $\cC$}
There are two main approaches in numerical construction of geodesic paths on manifolds. The first approach, called {\it path-straightening}, initializes with an arbitrary path between the given two points on the manifold and then iteratively ``straightens'' it until a geodesic is reached. The second approach, called the {\it shooting method}, 
tries to ``shoot"  a geodesic from the first point, iteratively adjusting the shooting direction, so that the resulting geodesic passes through the second point. In this paper, we use the shooting method to obtain the geodesic paths on $\cC$.

In order to implement the shooting method, we need the exponential map on $\cC$.  Given a point $(p,q) \in \cC$ and a tangent vector $(u,w) \in \T_{(p,q)}(\cC)$, the exponential map $\exp_{(p,q)}\left( s(u,w) \right)$ for $s \in [0,1]$  gives a geodesic path $(x(s),v(s))$ on $\cC$.  Equation \ref{gdequ1} helps us with this construction as
follows. The two equations essentially provide expressions for second-order covariant derivatives of $x$ and $v$ components of the path. Therefore, using numerical 
techniques, we can perform covariant integration of these quantities to recover the path itself.

\begin{algorithm} \label{algo2}
{ \bf Numerical exponential map on $\cC$}

Let the initial point  be $(x(0),v(0)) \in \cC$ and the tangent vector be $(u,w) \in \T_{(x(0),v(0))}(\cC)$. We have $x_s(0) = u$, $\nabla_{x_s} v(s)|_{s=0} = w$.  Fix an $\epsilon = 1/n$, the exponential map $(x(i\epsilon),v(i\epsilon)) = \exp_{(x(0),v(0))}\left( i \epsilon(u,w) \right)$ ($i=1,\cdots,n$) is given as:
\begin{enumerate}
\item Set $x(\epsilon) = \exp_{x(0)}(\epsilon x_s(0))$, where $x_s(0) = u$, and $v(\epsilon) = (v^{\parallel} + \epsilon w^{\parallel})$, where $v^{\parallel}$ and $w^{\parallel}$ are parallel transports of $v(0)$ and $w$ along path $x$ from $x(0)$ to $x(\epsilon)$, respectively. 

\item For each i = 1,2,...,n-1, calculate 
$$
x_s(i \epsilon) = \left[x_s ((i-1)\epsilon) + \epsilon \nabla_{x_s} x_s((i-1)\epsilon)\right ]_{x((i-1)\epsilon) \rightarrow x(i \epsilon)},
$$ 
where 
$
\nabla_{x_s} x_s((i-1)\epsilon) = -R \left( v((i-1)\epsilon), \nabla_{x_s} v((i-1)\epsilon) \right) \left( x_s( (i-1)\epsilon ) \right)
$
is given by the first equation in Theorem \ref{thm:geodesicequ}. 
It is easy to show that $R \left( v((i-1)\epsilon), \nabla_{x_s} v((i-1)\epsilon) \right) =  R\left( v^{\parallel} +\epsilon(i-1) w^{\parallel} ,w^{\parallel} \right) = R\left( v^{\parallel},w^{\parallel} \right)$, where $v^{\parallel} = v(0)_{x(0) \rightarrow x((i-1)\epsilon)}$, and $w ^{\parallel} = w_{x(0) \rightarrow x((i-1)\epsilon)}$.

\item Obtain $x((i+1)\epsilon) = \exp_{x\left(i\epsilon \right)} \left(\epsilon x_s(i\epsilon) \right)$, and $v((i+1)\epsilon) = v^{\parallel} + (i+1)\epsilon w^{\parallel}$, where  $v^{\parallel} = v(0)_{x(0) \rightarrow x((i+1)\epsilon)}$, and $w ^{\parallel} = w_{x(0) \rightarrow x((i+1)\epsilon)}$. 
\end{enumerate}
\end{algorithm}
Once we have a numerical procedure for the exponential map, 
we can establish the shooting method for finding geodesics. Let $(p_1,q_1)$ be the starting point and $(p_2,q_2)$ be the target point, the shooting method 
iteratively updates the tangent or shooting vector $(u,w)$ on $\T_{(p_1,q_1)}(\cC)$ such that $\exp_{(p_1,q_1)}\left((u,w)\right) = (p_2,q_2)$. 
Then, the geodesic between $(p_1,q_1)$ and $(p_2,q_2)$ is given by $(x(s),v(s))=\exp_{(p_1,q_1)}(s(u,w))$, $s\in[0,1]$.  The key step here is to use
the current discrepancy between the point reached, $\exp_{(p_1,q_1)}\left((u,w)\right)$, and the target, $(p_2, q_2)$, to update
the shooting vector $(u,w)$, at each iteration. There are several possibilities for performing 
the updates and we discuss one  here. Since we have two components to update, $u$ and $w$, we will update them 
separately: (1) Fix $w$ and update $u$. For the $u$ component, the increment can come from parallel translation of the vector $\exp_{\tilde{p}}^{-1}(p_2)$ (the difference between the reached point $\tilde{p}$ and the target point $p_2$) from $\tilde{p}$ to $p_1$, where
$\tilde{p}$ is the first component of reached point $\exp_{(p_1,q_1)}((u,w))$. (2) Fix $u$ and update $w$. For the $w$ component, we can take the difference between 
$q_2$ and the second component of the point reached (denoted as $\tilde{q}$) as the increment. This is done by parallel translating $\tilde{q}$ to $\T_{p_2}(M)$ (the same space as $q_2$) and calculate the difference, and then parallel translate the difference to $\T_{p_1}(M)$ to update $w$.
  
\begin{algorithm} \label{algo:shoot}
{\bf Shooting algorithm for calculating geodesic on $\cC$}

Given $(p_1,q_2),(p_2,q_2) \in \cC$, select one point, say $(p_1,q_1)$, as the starting point and the other, $(p_2,q_2)$, as the target point. the shooting algorithm for calculating the geodesic from $(p_1,q_1)$ to $(p_2,q_2)$ is:
\begin{enumerate}
\item Initialize the shooting direction: find the tangent vector $u$ at $p_1$ such that the exponential map $\exp_{p_1}(u) = p_2$ on the manifold $M$. Parallel transport $q_2$ to the tangent space of $p_1$ along the shortest geodesic between $p_1$ and $p_2$, denoted as $q^{\parallel}_2$. Initialize $w = q^{\parallel}_2 - p_1$. Now we have a pair $(u,w) \in \T_{(p_1,q_1)}(\cC)$.
\item Construct a geodesic starting from $(p_1,q_1)$ in the direction $(u,w)$ using the numerical exponential map in Algorithm \ref{algo2}. Let us denote this geodesic path as $(x(s),v(s))$, where $s$ is the time parameter for the geodesic flow. 
\item If $ (x(1),v(1))  = (p_2,q_2)$, we are done. If not, measure the discrepancy between $(x(1),v(1))$ and $(p_2,q_2)$ using a simple measure, e.g. $\ltwo$ distance. 
\item Iteratively, update the shooting direction $(u,w)$ to reduce the discrepancy to zero. This update can be done using a two-stage approach: (1) fix $u$ and update $w$ until converge; (2) fix $w$ and update $u$ until converge. 
\end{enumerate} 
\end{algorithm}

Recall that trajectories on $M$ and their representations in $\cC$ are bijective. For each pair $(p,q) \in \cC$, one can reconstruct the 
corresponding trajectory $\alpha$ using covariant integration. 
 A numerical implementation of this procedure is summarized in Algorithm \ref{algo1}. 

\begin{algorithm}\label{algo1}
{\bf Covariant integral of $q$ along $\alpha$} 

 Given a TSRVF $q$ sampled at $T$ times $\{ t \delta | t=0,1, \dots, T-1\}, \delta= 1/T$, 
and the starting point $p$: 
\begin{enumerate}
\item Set $\alpha(0)=p$, and compute $\alpha(\delta)=\exp_{\alpha(0)}(\delta q(0) |q(0)|)$, where $\exp$ denotes the exponential map on $M$.
\item For $t=1, 2, \dots, T-1$
\begin{enumerate}
\item Parallel transport $q(t\delta)$ to $\alpha(t\delta)$ along the current trajectory from $\alpha(0)$ to $\alpha(t\delta)$, and call it $q^{\parallel}(t\delta)$. 
\item Compute 
$
\alpha((t+1)\delta)=\exp_{\alpha(t\delta)}(\delta q^{\parallel}(t\delta)   |q^{\parallel}(t\delta)|) .
$
\end{enumerate}
\end{enumerate}
\end{algorithm}

\begin{figure}
\begin{center}
\begin{tabular}{c}
\includegraphics[height=2.2in]{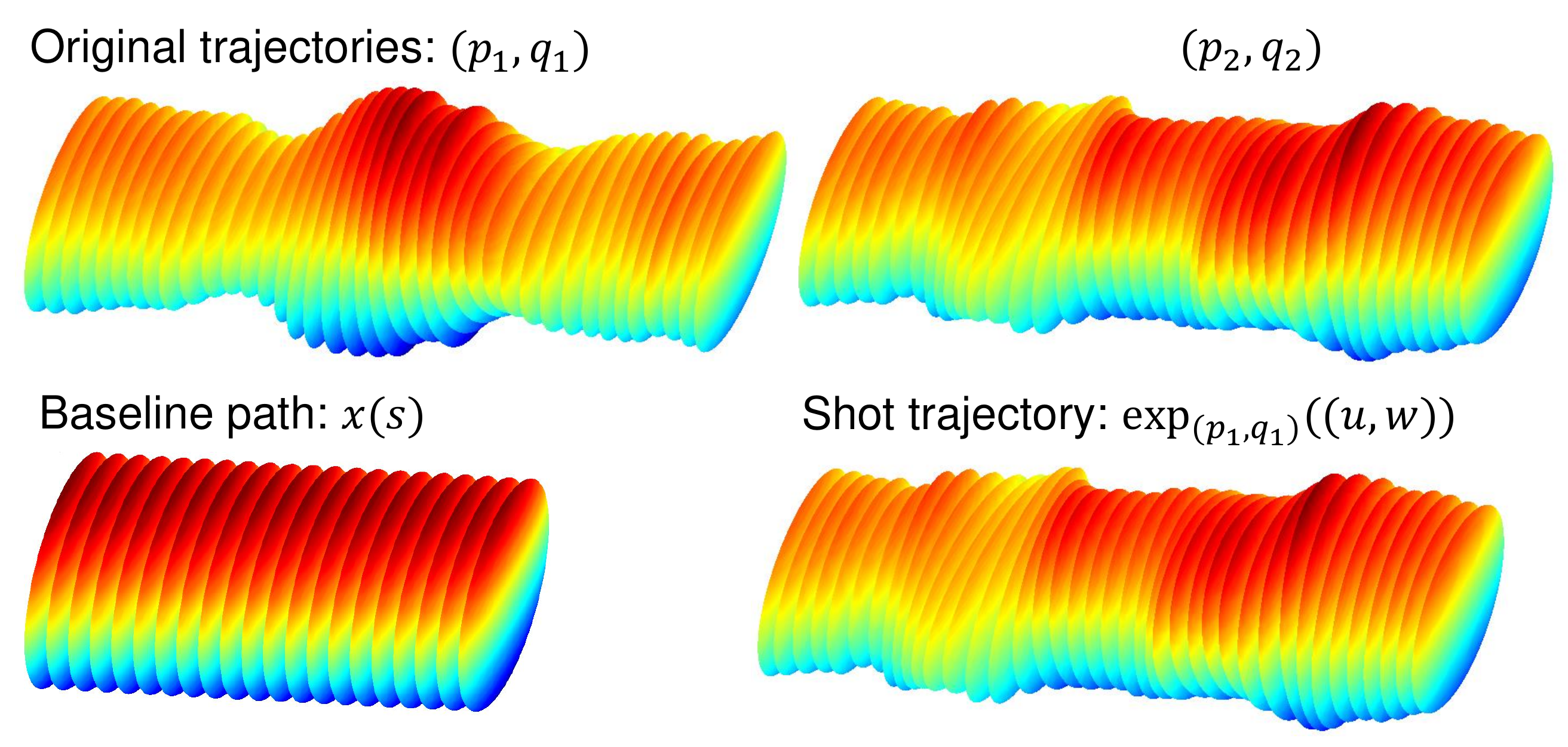}\\
\includegraphics[height=1.4in]{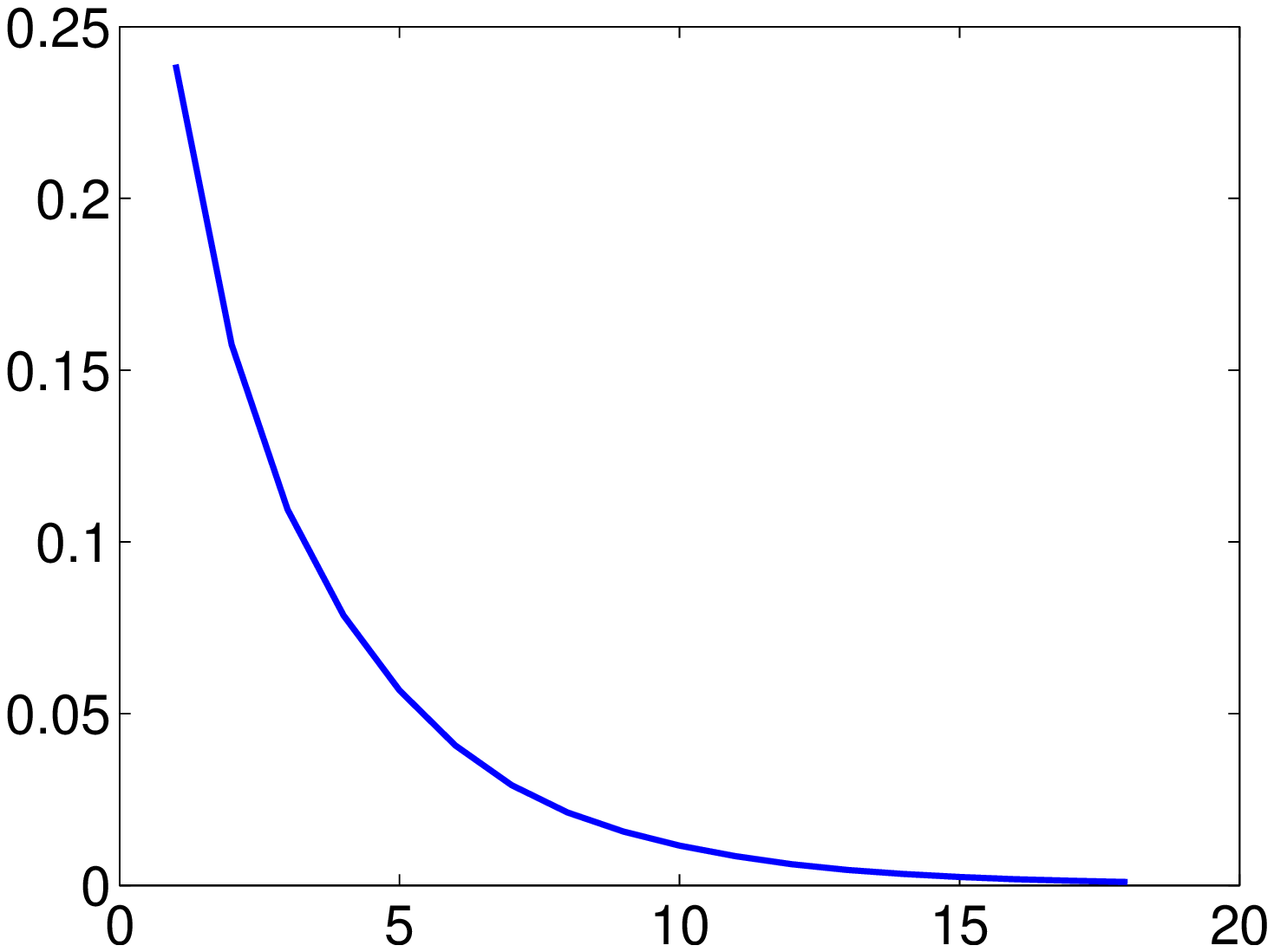}\\
\end{tabular}
\caption{Example of calculating geodesic using shooting method for trajectories on $\P$. 
} \label{fig:spd1geodesc} 
\end{center}
\end{figure}

Algorithm \ref{algo:shoot} allows us to calculate the geodesic between two points in $\cC$. 
So, for each point along the geodesic $(x(s),v(s))$ in $\cC$, one can easily reconstruct the trajectory 
 on $M$ using Algorithm \ref{algo1}.
Here, one sets $x(s)$ as the starting point and $v(s)$ as the TSRVF of the trajectory. Fig. \ref{fig:spd1geodesc} shows one example of calculating geodesic using the numerical method in Algorithm \ref{algo:shoot}, where $M = \P$,  the set of $3 \times 3$ SPDMs. In this plot, each SPDM matrix is visualized by an ellipsoid
and a trajectory on $\tilde{\P}$ by a sequence of 
ellipsoids.  The left panel shows two original trajectories $\alpha_1$ and $\alpha_2$ (their representations are $(p_1,q_1)$ and $(p_2,q_2)$), the 
end point of the trajectory shot ($\exp_{(p_1,q_1)}(u,w)$),  and baseline path $x(s)$. In this case, we selected $(p_1,q_1)$ as the starting point and computed the shooting direction $(u,w)$ such that $\exp_{(p_1,q_1)}(u,w) = (p_2,q_2)$. The bottom panel shows the 
evolution of $\ltwo$ norm between the shot trajectory and the target $(p_2,q_2)$ during the shooting algorithm. 

\subsection{Geodesic Distance on $\cC$}
Using the natural Riemannian metric on $\cC$ (defined in Eqn. \ref{eqn:ccrp}), the geodesic distance between the two points is defined as the following.  
\begin{defn} \label{defn:geod}
Given two trajectories $\alpha_1$, $\alpha_2$ and their representations $(p_1,q_1), (p_2,q_2) \in \cC$, and let $(x(s),v(s)) \in \cC$, $s \in [0,1]$ be the geodesic between $(p_1,q_1)$ and  $(p_2,q_2)$ on $\cC$, 
the geodesic distance is given as:
\begin{equation}\label{geodistance}
d_c((p_1,q_1), (p_2,q_2))=\sqrt{ l_{x}^2 + \int_0^1 |{ q}^{\parallel}_1(\tau)- q_2(\tau)|^2d\tau } \ \ .
\end{equation}
\end{defn}
This distance has two components: (1) the length between the starting points on $M$, $l_x = \int_0^1 |\dot{x}(s) |ds$; and (2) the standard $\ltwo$ norm on $\cC_{p_2}$ between the TSRVFs of the two trajectories, where ${ q}^{\parallel}_1$ represents the parallel transport of $q_1\in\cC_{p_1}$ along $x$ to $\cC_{p_2}$. 
Since we have a numerical approach for approximating the geodesic, this same algorithm can also provide an estimate for the geodesic distance.

\subsection{Geodesic Distance on Quotient Space $\cC/\Gamma$}
The main motivation of using TSRVF representation for trajectories on $M$ and constructing the distance $d_c$ to compare two trajectories comes from the following.
If a trajectory $\alpha$ is warped by $\gamma$, resulting in $\alpha \circ \gamma$, what is the TSRVF
of the time-warped trajectory? The new TSRVF is given by:
\begin{eqnarray*}
q_{\alpha \circ \gamma}(t) &=& \left({  (\dot{\alpha}(\gamma(t)) \dot{\gamma}(t)) \over \sqrt{ |  \dot{\alpha}(\gamma(t)) \dot{\gamma}(t)|} } \right)_{\alpha(\gamma(t)) \rightarrow p}
=
\left({  (\dot{\alpha}(\gamma(t)))\sqrt{\dot{\gamma}(t)}  \over \sqrt{ |  \dot{\alpha}(\gamma(t)) |} } \right)_{\alpha(\gamma(t)) \rightarrow p}  \\
&=& q_\alpha(\gamma(t)) \sqrt{\dot{\gamma}(t)} \equiv (q_\alpha*\gamma)(t)\ .
\end{eqnarray*}

\begin{thm} \label{thm:isometric}
For any two trajectories $\alpha_1, \alpha_2 \in {\cal F}$ and  their representations $(p_1,q_{1}), (p_2,q_{2}) \in \cC$, the metric
$d_c$ satisfies $d_c((p_1,q_{\alpha_1\circ \gamma}), (p_2,q_{\alpha_2 \circ \gamma})) = d_c((p_1,q_{1}), (p_2,q_{2}))$, for any $\gamma \in \Gamma$ 
\end{thm}
\noindent {\bf Proof}: 
First, if a trajectory is warped by $\gamma \in \Gamma$, the resulting trajectory is $\alpha \circ \gamma$, and the starting point does not change. If the original representation of $\alpha$ is $(p,q) \in \cC$, the resulting representation is $(p,q_{\alpha \circ \gamma})$.
Second, the baseline $x(s)$ connecting the starting points of given two trajectories will not change with respect to arbitrary $\gamma$'s since the starting points are fixed, so for $l_{x}$. 
Starting with the left side, we simplify 
$d_c^2( (p_1,q_{\alpha_1\circ \gamma}), (p_2,q_{\alpha_2 \circ \gamma}) )$ as follows:
\begin{equation}
\begin{array}{lcl}
 &=&   l_{x}^2 +   \int_0^1 | ({q}^{\parallel}_{\alpha_1}*\gamma)(t) -
(q_{\alpha_2}*\gamma)(t) |^2 dt\\
&  = & l_{x}^2+ \int_0^1 | {q}^{\parallel}_{\alpha_1}(\gamma(t)) \sqrt{\dot{\gamma}(t)} - q_{\alpha_2}(\gamma(t)) \sqrt{\dot{\gamma}(t)}  |^2 dt \\
&=& l_{x}^2 +  \int_0^1 | { q}^{\parallel}_{\alpha_1}(\gamma(t)) -
q_{\alpha_2}(\gamma(t))  |^2  \dot{\gamma}(t)~dt  \\
&=&  l_{x}^2+  \int_0^1 | { q}^{\parallel}_{1}(s) -
q_{2}(s)  |^2 ~ds = d_c^2((p_1,q_{1}), (p_2,q_{2}))\ ,
\end{array}
\end{equation}
where we have used $s = \gamma(t).$

Theorem  \ref{thm:isometric} reveals the advantage of using TSRVF representation: the action of $\Gamma$ on $\cC$ under the metric $d_c$ is by isometries.  The isometry property of time-warping action under the metric $d_c$ allows us to compare trajectories in a manner that the comparison is invariant to the time warping. This is achieved through defining a distance in the quotient space of reparameterization group. 

To form the quotient space of $\cC$ modulo the re-parameterization group, we first introduce 
$\tilde{\Gamma}$ as the set of all non-decreasing, absolutely continuous functions $\gamma$ on $[0,1]$
such that $\gamma(0) = 0$ and $\gamma(1) = 1$. This set is a semigroup with the composition operation
(it does not have a well-defined inverse). It can be shown that $\Gamma$ is a dense subset of $\tilde{\Gamma}$. 
Consequently, the orbit of a TSRVF $q$ under the action of $\tilde{\Gamma}$ is exactly the same 
as the closure of the orbit of $q$ under the action of $\Gamma$. Since the orbits under $\tilde{\Gamma}$ 
form closed sets \cite{su2014}, while those under $\Gamma$ do not, we choose to work with the former, at least for the 
formal development. But in practice, we will approximate the theoretical solutions using the elements of $\Gamma$. We define the quotient space 
$\cC/\tilde{\Gamma}$ as the set of all orbits under the action of $\tilde{\Gamma}$, with each orbit being:
$$
[(p,q)] \equiv (p,[q]) = \{ (p,(q\circ\gamma)\sqrt{\dot{\gamma}})| \gamma \in \tilde{\Gamma} \} \ .
$$
To understand the orbit, one can view it as an equivalence class. For any two trajectories $\alpha_1,\alpha_2$ and their representations in $\cC$, $(p_1,q_{1}),(p_2,q_{2})$, we define them to be equivalent when: (1) $p_1 = p_2$; and (2) there exists a sequence $\gamma_i \in \tilde{\Gamma}$ such that $q_{\alpha_2 \circ \gamma_i}$ 
converges to $q_{1}$. In other words, if two trajectories have the same starting point, and the TSRVF of one can be time-warped into the TSRVF of the other, 
using a sequence of time-warpings, then these two trajectories are deemed equivalent to each other. 
Theorem \ref{thm:isometric} indicates that if two trajectories are warped by the same $\gamma$ function, the distance $d_c$ between them remains the same. In other word, the orbits in $\cC$ are ``parallel" to each other. 

Our goal is to define a distance such that it is invariant to the time-warping of trajectories. This can be achieved by comparing trajectories through their equivalence classes (or the orbits). That is, define a metric on the quotient space $\cC/\tilde{\Gamma}$ using the inherent Riemannian metric from $\cC$. The geodesic distance on $\cC/\tilde{\Gamma}$ is defined as follows. 

\begin{defn} \label{defn:geoQuo}
       The geodesic distance $d_{q}$ on $\cC/\tilde{\Gamma}$ is the shortest distance between two orbits in $\cC$, given as	
\begin{eqnarray}\label{geodistq}
&&d_q((p_1,[q_1]), (p_2,[q_2])) \nonumber \\
&=& \inf_{\gamma_1, \gamma_2 \in \tilde{\Gamma}}d_c( (p_1,(q_1 \circ \gamma_1)\sqrt{\dot{\gamma}_1})), (p_2,(q_2 \circ \gamma_2)\sqrt{\dot{\gamma}_2}) )    \nonumber\\
 &\approx&  \inf_{\gamma \in \Gamma}d_c( (p_1,q_1), (p_2,(q_2 \circ \gamma)\sqrt{\dot{\gamma}}) ) \ \ .
\end{eqnarray}
\end{defn}

The geodesic $d_q$ between $(p_1,[q_1])$ and $(p_2,[q_2])$ is obtained by forming geodesics between all possible cross pairs in sets $(p_1,[q_1])$ and $(p_2,[q_2])$. 
Since the group action is by isometries, we can fixed one point, say $(p_1,q_1)$, and search over all $(p_2,[q_2])$ that minimizes Eqn. \ref{geodistq}. 

Next, we focus on the problem of pairwise temporal registration between two trajectories. 
 Eqn. \ref{geodistq} not only defines a metric on the quotient space $\cC/\tilde{\Gamma}$ but also provides an objective function for registering two trajectories: the optimal $\gamma^*$ for Eqn. \ref{geodistq} aligns $\alpha_2$ to $\alpha_1$. That is the point $\alpha_1(t)$ is optimally matched with the point $\alpha_2(\gamma^*(t))$. To solve Eqn. \ref{geodistq}, it is equivalent to optimize the following equation:
\begin{equation}\label{eqn:pairwish}
\min_{(x,v), \gamma}\left( {\ l^2_{x} + \int_0^1\|{q}^{\parallel}_{1,x}(t)- (q_2, \gamma(t))\|^2dt\ }     \right) \quad,
\end{equation}
where $(x,v)$ is the geodesic between $(p_1,q_1)$ and $(p_2,q_{\alpha_2 \circ \gamma})$, and ${q}^{\parallel}_{1,x}$ means parallel transport $q_1$ along $x$ to $\cC_{p_2}$.
 Note that the time-warping $\gamma$ acting on $\alpha_2$ changes the underlying geodesic $(x,v)$ between two trajectories. Algorithm \ref{algo:pairwise} describes a numerical solution for optimizing Eqn. \ref{eqn:pairwish} on a general manifold $M$. 

\begin{algorithm}\label{algo:pairwise}
{\bf Pairwise registration of two trajectories on $M$}

Represent two trajectories $\alpha_1(t),\alpha_2(t)$ by their TSRVFs, $(p_1,q_1)$ and $(p_2,q_2)$. Let $\gamma^* = \gamma_{id}$, and set $itermax = K$ (a large integer), $iter = 1$ and a small $\epsilon>0$.

\begin{enumerate}
\item Select one point, say $(p_1,q_1)$, as the starting point and the other, $(p_2,\tilde{q}_2)$, as the target point, where $\tilde{q}_2$ denotes $(q_{\alpha_2}*\gamma)$, for $\gamma \in \Gamma$. In this step, let $\gamma = \gamma_{id}$.

\item  Obtain $(u,w) \in \T_{(p_1,q_1)}(\cC)$ such that $\exp_{(p_1,q_1)}(s(u,w)) = (x(s),v(s)),s \in [0,1]$ and $(x(1),v(1))=(p_2,\tilde{q}_2)$.

\item Parallel transport $\tilde{q}_2$ to the tangent space $\T_{p_1}(M)$ along $x(s)$, denoted as $\tilde{q}_2^{\parallel}$. Align $\tilde{q}_2^{\parallel}$ to $q_1$ using Dynamic Programming Algorithm and obtain the optimal warping function $\gamma$. 

\item Update $\gamma^* =\gamma^*\circ \gamma $ by composition. 
If $\|\gamma - \gamma_{id}\| < \epsilon$ or $iter>itermax$ stop. Else, set $\tilde{q}_2 = (\tilde{q}_{\alpha_2}*\gamma)$, $iter = iter +1$ and go back to step 3. 
\end{enumerate}
\end{algorithm}

\begin{figure}[]
\begin{center}
\begin{tabular}{cc}
\includegraphics[height=1.8in]{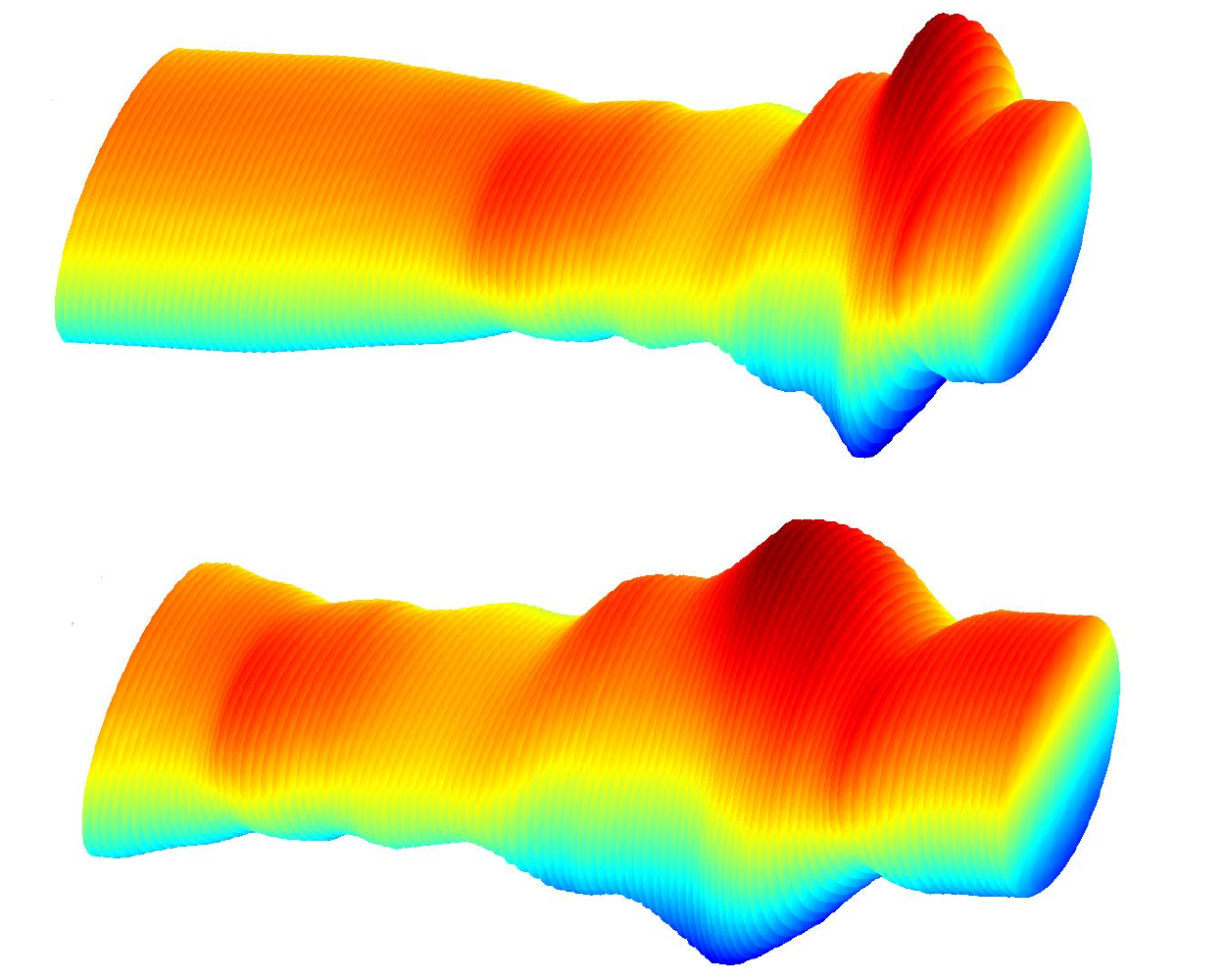}&
\includegraphics[height=1.8in]{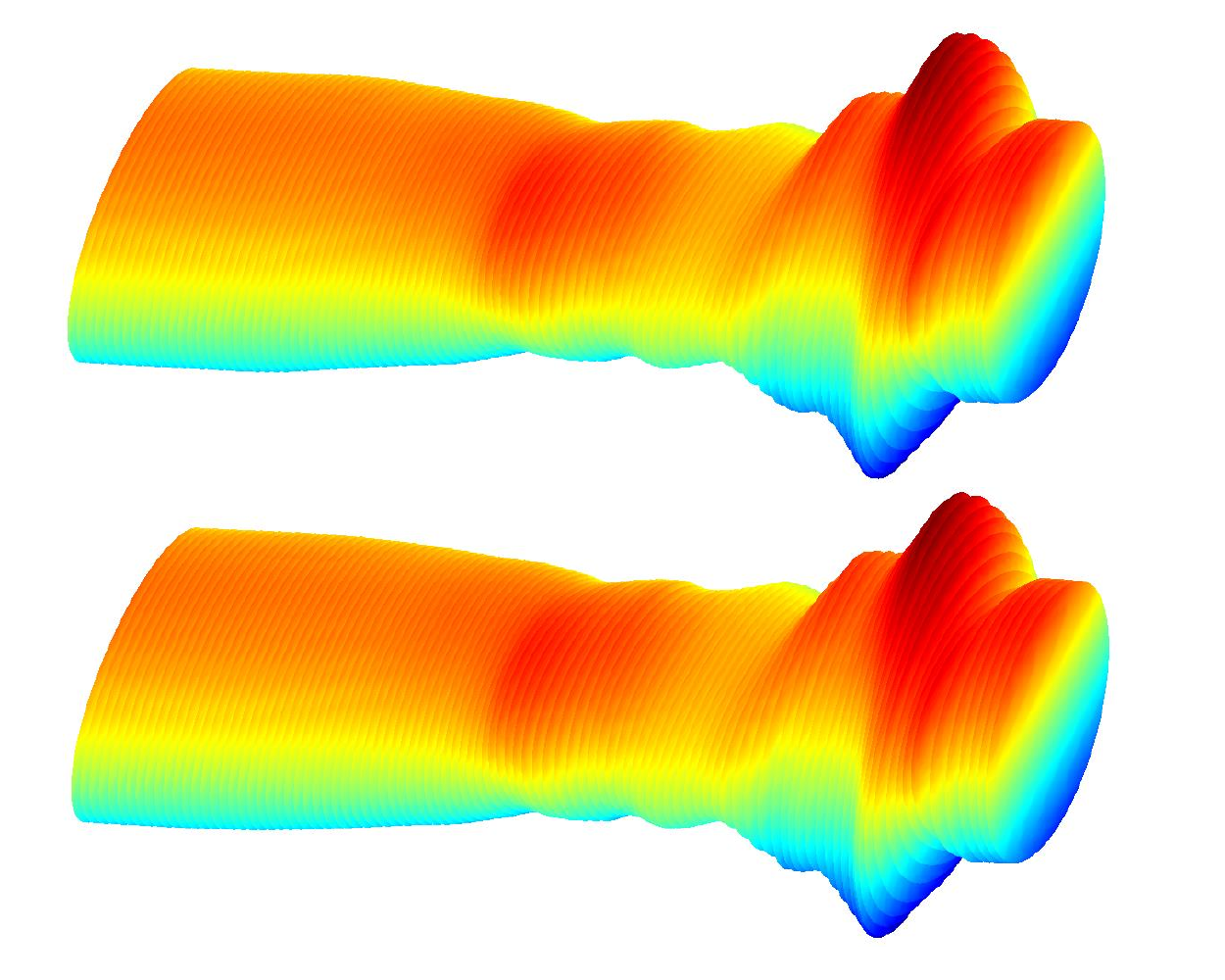}\\
Before: $\alpha_1$ and $\alpha_2$  & After: $\alpha_1$ and $\alpha_2\circ\gamma^* $ \\

\multicolumn{2}{c}{\includegraphics[height=1.4in]{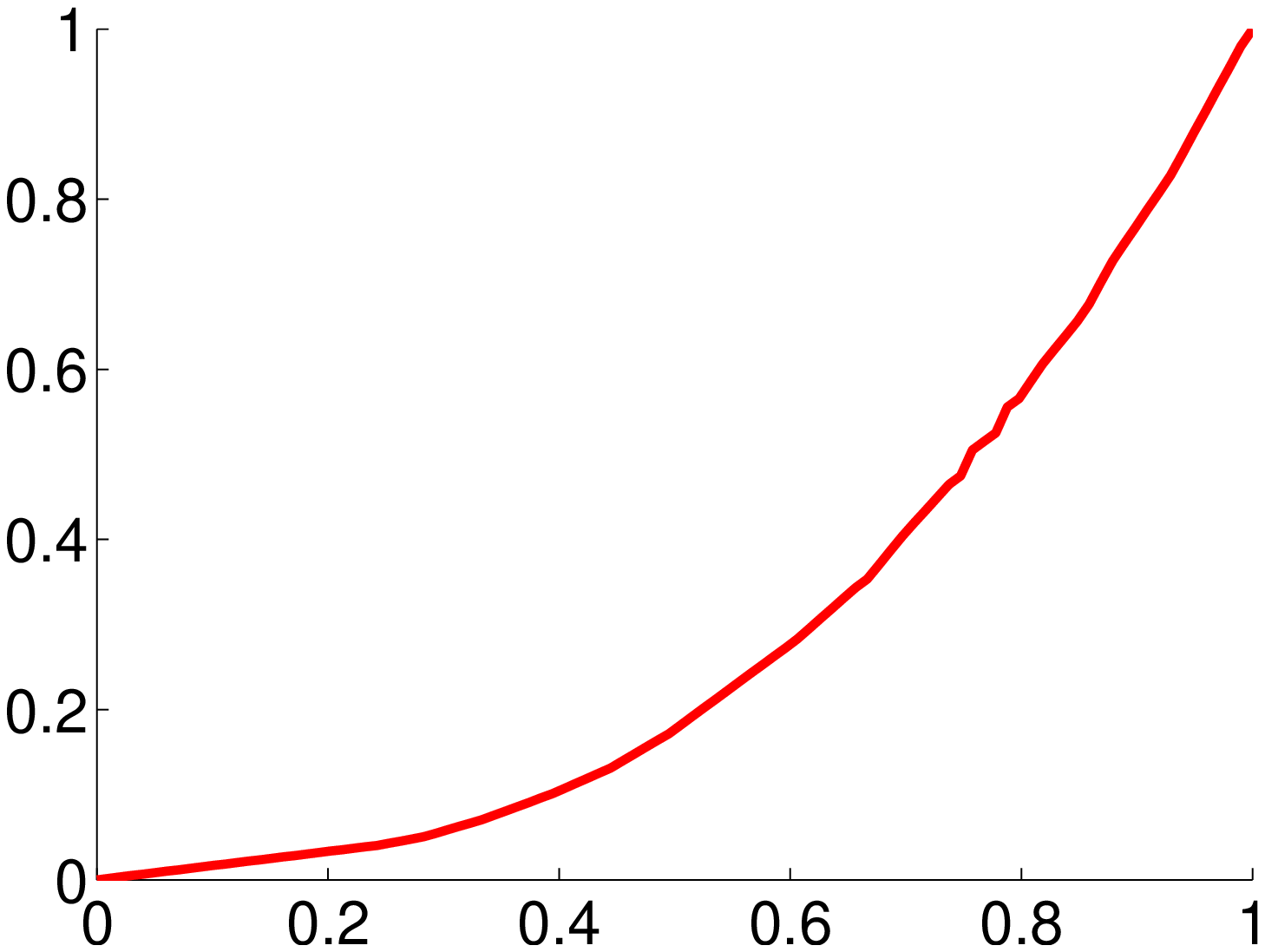}}\\
\multicolumn{2}{c}{$\gamma^*$}\\
\end{tabular}
\caption{Pairwise registration of two trajectories. }\label{fig:alignment} 
\end{center}
\end{figure}

Note that Step 2 corresponds to the first argument $(x,v)$ and Step 3 corresponds to the second argument $\gamma$ in Eqn. \ref{eqn:pairwish}, respectively. The
optimization over the warping function in Step 3 is achieved using the Dynamic Programming Algorithm \cite{BertsekasDP01}. 
Here one samples the interval $[0, 1]$ using $N$ discrete points and then restricts to
only piecewise linear $\gamma$'s that pass through that $N \times N$ grid. 
In Fig. \ref{fig:alignment}, we present one example of aligning two trajectories $\alpha_1$ and $\alpha_2$ on space of $3 \times 3$ SPDMs. \\

\noindent {\bf Fast Approximation}:
Since pairwise temporal registration algorithm involves multiple evaluations of the exponential map and dynamic programming
alignment, it is not computationally efficient. One way to speed up this optimization is to use an approximate method: approximate the baseline $x(s)$ connecting two trajectories first (using geodesic between the starting points of trajectories) and then align their TSRVFs. Another way is to find an explicit expression for the exponential map. This 
seems possible only for a simple manifold, such as $M=\s^2$, but for a complicated manifold, such as $M=\tilde{\P}$, the 
analytical expressions are not known. In the experiment section, we use the approximate method to speed up the registration and comparison.

If we compare the proposed framework with that in  \cite{su2014}, we see several advantages. The proposed work perseveres the invariance
properties achieved in \cite{su2014}, but does not require choosing a reference point. 
Also, the proposed framework naturally includes the difference between the starting points of two trajectories that are
ignored in  \cite{su2014}. Since the velocity vectors here are transported to the starting point of 
a trajectory, along that trajectory, as opposed to a transport to an arbitrary  reference point in  \cite{su2014}, this 
representation is more stable.

\subsection{Statistical Summarization of Multiple Trajectories}
Since $d_q$ defines a metric in the quotient space $\cC/\Gamma$, this framework allows us to perform statistical analysis of multiple trajectories in $\cC/\Gamma$. Given a set of trajectories $\{\alpha_i, \ i= 1\dots k\}$, we are interested in computing the average of these trajectories and using it as a template for registering these trajectories. This sample average is calculated using the notion of Karcher mean \cite{CPA:Karcher}. In the space of $\cC/\Gamma$, the Karcher mean is defined to be:
$$(\mu_p,[\mu_q]) = \argmin_{(p,[q]) \in \cC/\Gamma} \sum_{i=1}^n d_q ( (p,[q]), (p_i, [q_{\alpha_i}]) )^2\ .
$$
Note that $(\mu_p, [\mu_q])$ is an orbit (equivalence class of trajectories) and one can select any element of this orbit as a template to help to align multiple trajectories. 



\begin{algorithm}\label{algo:karchermean}
{\bf Karcher mean}

For each $\alpha_i$, compute its TSRVF $q_i$, denote as $(p_i,q_i)$. Let $(\mu_p^j,\mu_q^j)$, $j=0$ be the initial estimate of the Karcher mean (e.g. 
we can choose one of the trajectories). Set small $\epsilon, \epsilon_1, \epsilon_2>0$.
\begin{enumerate}
\item For $i=1$ to $n$, align each trajectory $(p_i,q_i)$ to $(\mu_p^j,\mu_q^j)$ according to Algorithm \ref{algo:pairwise}, denoted as $(p_i,\tilde{q}_i)$. Algorithm \ref{algo:pairwise} also gives us the inverse exponential map: $(u_i, w_i) = \exp^{-1}_{(\mu_p^j,\mu_q^j)}(p_i,\tilde{q}_i)$. 

\item Compute the average direction: $\bar{u} = \frac{1}{n}\sum_{i=1}^n u_i$, $\bar{w} = \frac{1}{n}\sum_{i=1}^n w_i$.

\item If $||\bar{u}||<\epsilon_1$  and $ || \bar{w} ||<\epsilon_2$, stop. Else, update $(\mu_p^j,\mu_q^j)$ in the direction of $(\bar{u},\bar{w})$ using exponential map:
$ (\mu_p^{j+1},\mu_q^{j+1}) = \exp_{(\mu_p^j,\mu_q^j)}(\epsilon \bar{u}, \epsilon \bar{w}) $,
where $\epsilon$ is the step size. We often let $\epsilon=0.5$.

\item Set $j=j+1$, return to step 2. 
\end{enumerate}
\end{algorithm}

\begin{figure}[]
\begin{center}
\begin{tabular}{|cc|}
\hline
\multicolumn{2}{|l|}{Simulated trajectories:}\\
\includegraphics[height=1.8in]{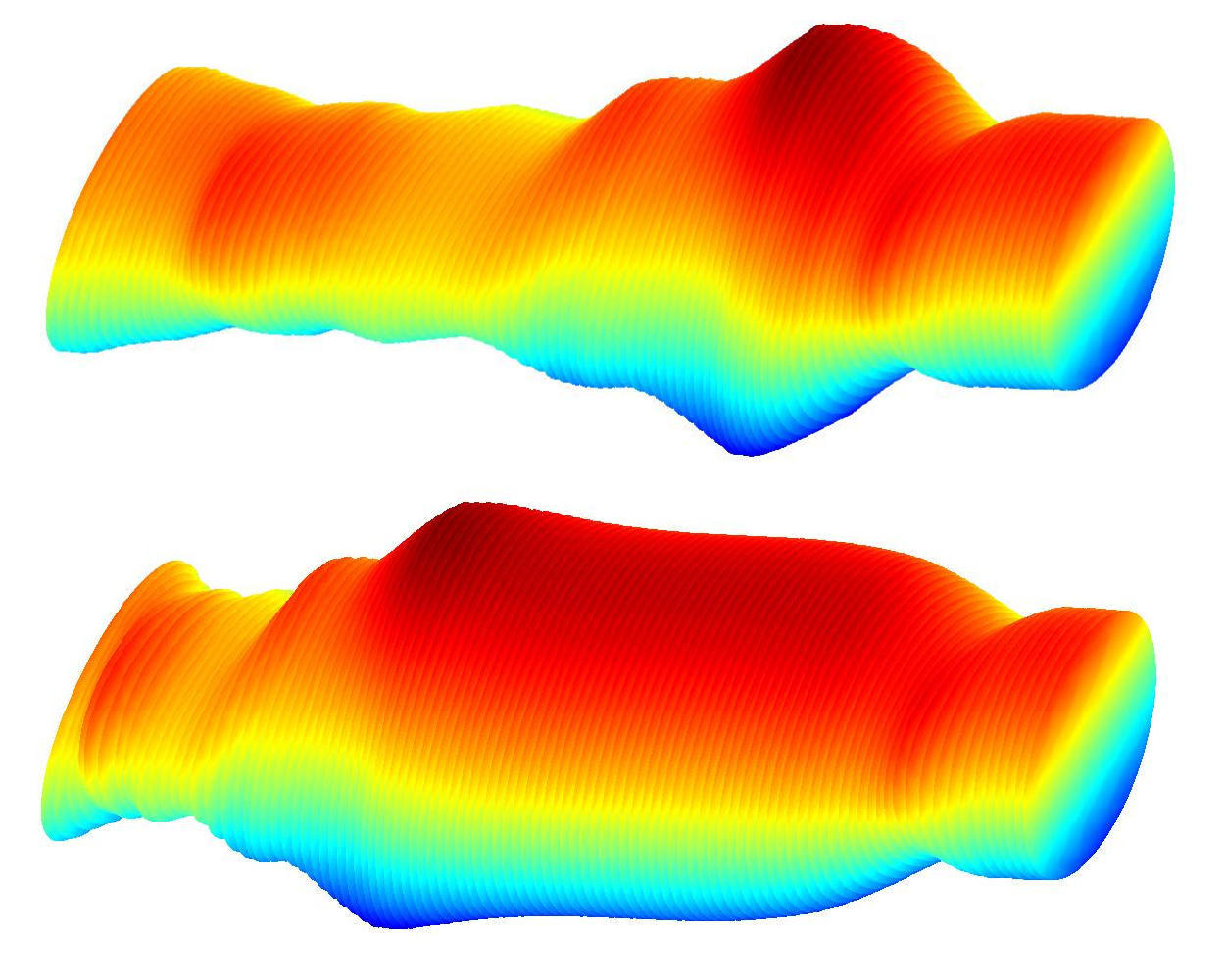}&
\includegraphics[height=1.8in]{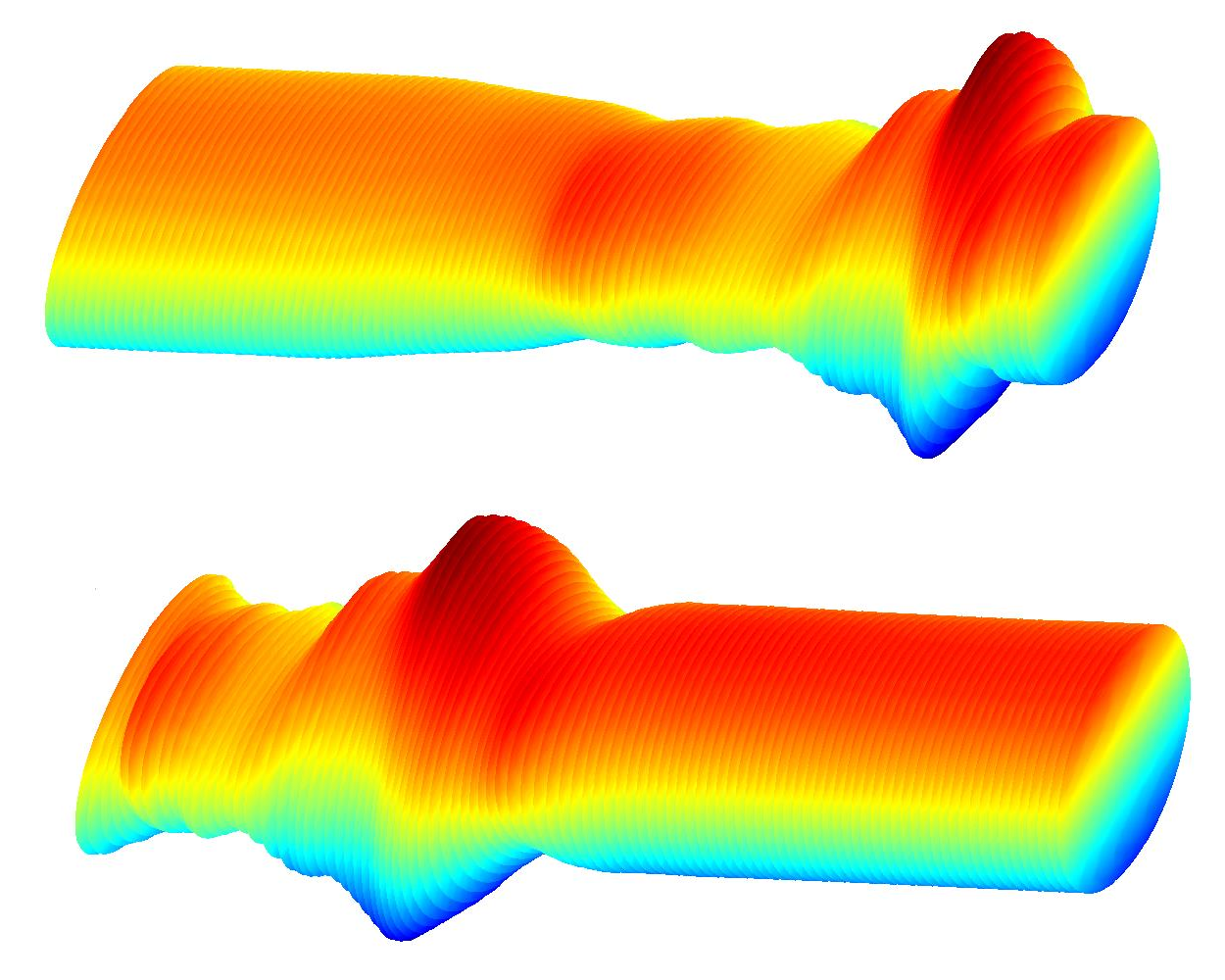} \\
\hline
\end{tabular}
\begin{tabular}{|cc|}
\hline
\includegraphics[height=0.90in]{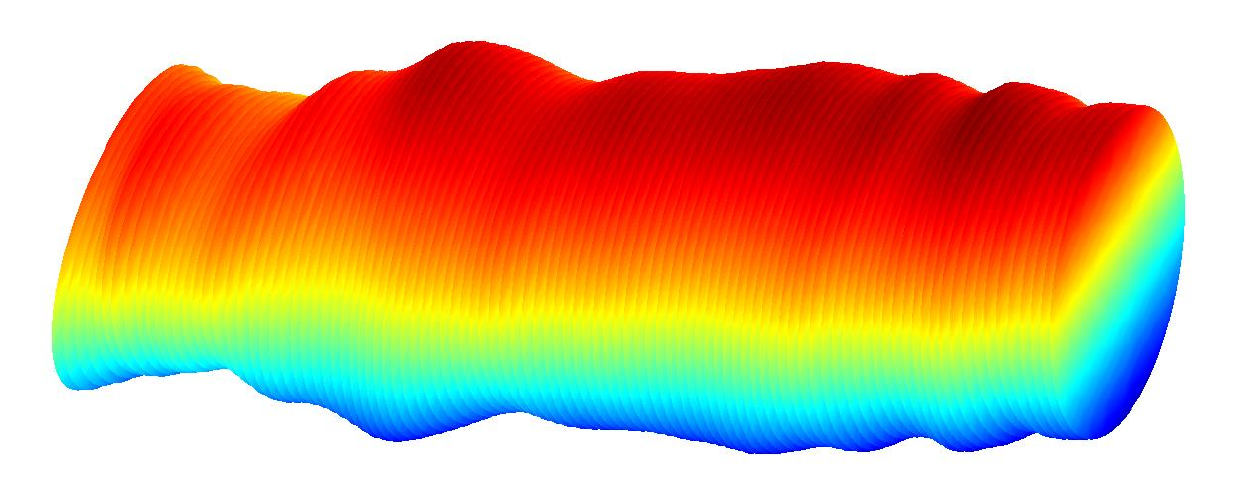}&
\includegraphics[height=0.90in]{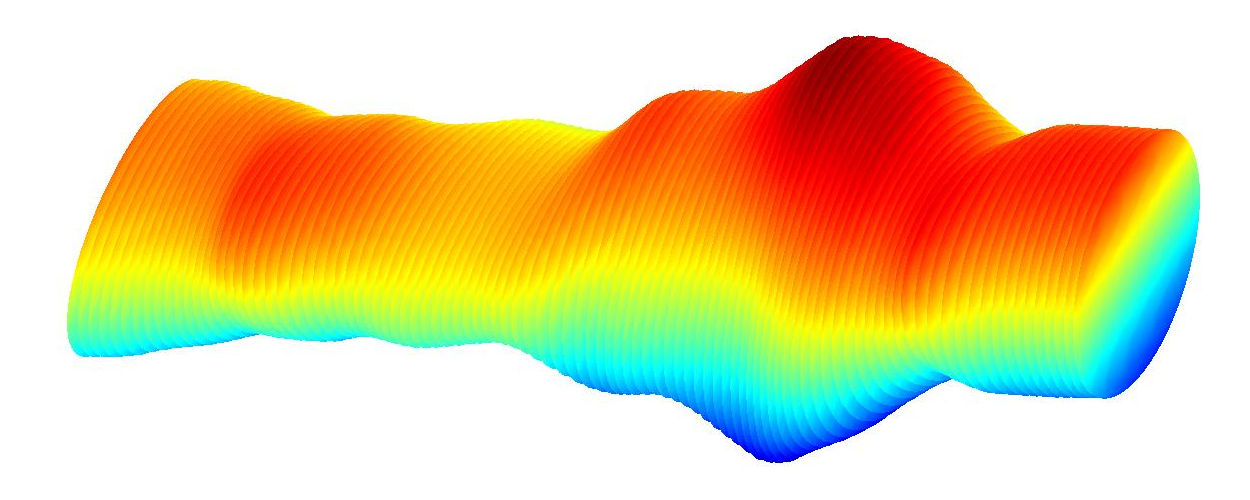}\\
Mean before registration & Mean after registration \\
\hline
\end{tabular}
\caption{Example of calculating the mean trajectory. The upper panel shows simulated trajectories, and the bottom panel shows mean before and after alignment. }\label{fig:mean} 
\end{center}
\end{figure}

After obtaining the converged $(\mu_p,\mu_q)$, one can compute the covariant integral using Algorithm \ref{algo1}, denoted by $\mu$, which is the Karcher mean of $\{\alpha_1,\alpha_2,...,\alpha_n\}$. Fig. \ref{fig:mean} shows one result on calculating the mean of given simulated trajectories. The upper panel shows the simulated trajectories. The bottom panel shows mean trajectory in two cases: before alignment and after alignment. One can see that after the alignment, the structure of these trajectories are preserved. 

\section{Video-Based Activities Recognition} \label{sec:exp}
Now we turn to evaluation of the framework developed so far on test datasets involving action recognition using videos.  While there are several 
application secures involving video-based action recognition, we try two applications here: (1) hand-gesture recognition, where one classifies the 
hand motion into pre-determined classes using video data, and (2)  visual speech recognition, 
where one classifies phrases uttered using videos of lip movements.

\subsection{Representations of Videos as Trajectories on $\tilde{\P}$}
The video data is typically extremely high-dimensional, a short video with 50-frames could have more than a million pixel values. 
To represent a video or a frame in this video, researchers extract few relevant features from the data, forming low-dimensional 
representations and devise statistical 
analysis of these features directly. 
One common technique is to extract a covariance matrix of relevant features to represent each 
frame in the video \cite{Tuzel2006,Porikli2006,Guo:2010}. The definition of a covariance feature for a frame is as follows: 
At each pixel (or patch) of the frame image, extract an $d$-dimension local feature (e.g. pixel location, intensity, spatial derivatives, HOG features, etc), and then calculate a covariance matrix of those local features over the whole frame, i.e. sum over the spatial coordinates. 
More precisely , let  $f_x \in \real^d$ denote a $d$-dimensional  feature at location $x$ in the image $I$.  The empirical estimate of the covariance matrix of $S$ is given by: 
$ P := \frac{1}{|I|}\sum_{x\in I} (f_x - \bar{f}) (f_x - \bar{f})^T$,
where $\bar{f} = \frac{1}{|I|} \sum_{x \in I} f_x$ is the empirical mean feature vector. The covariance matrix provides a natural way to fuse multiple local features, 
across the whole image. The dimension of the representation space, $(d^2 + d)/2$, is much smaller than the image size $|I|$.

A video of human activities is now replaced by a sequence of covariance matrices. 
Therefore, each video is represented 
by a parameterized trajectory on the space of $\tilde{\P}$, and we can utilize the previous
framework to analyze these trajectories. 
Sometime, as in the hand-gesture recognition, we can localize features better by dividing the image frame into 
four quadrants and computing a covariance matrix for each quadrant. Then, each frame is represented as an element of $\tilde{\P}^4$, and the whole
video as $t \mapsto  \alpha(t) \in \tilde{\P}^4$. 

\subsection{Hand Gesture Recognition}
\begin{figure}
\begin{center}
\begin{tabular}{c}
\includegraphics[height=3in]{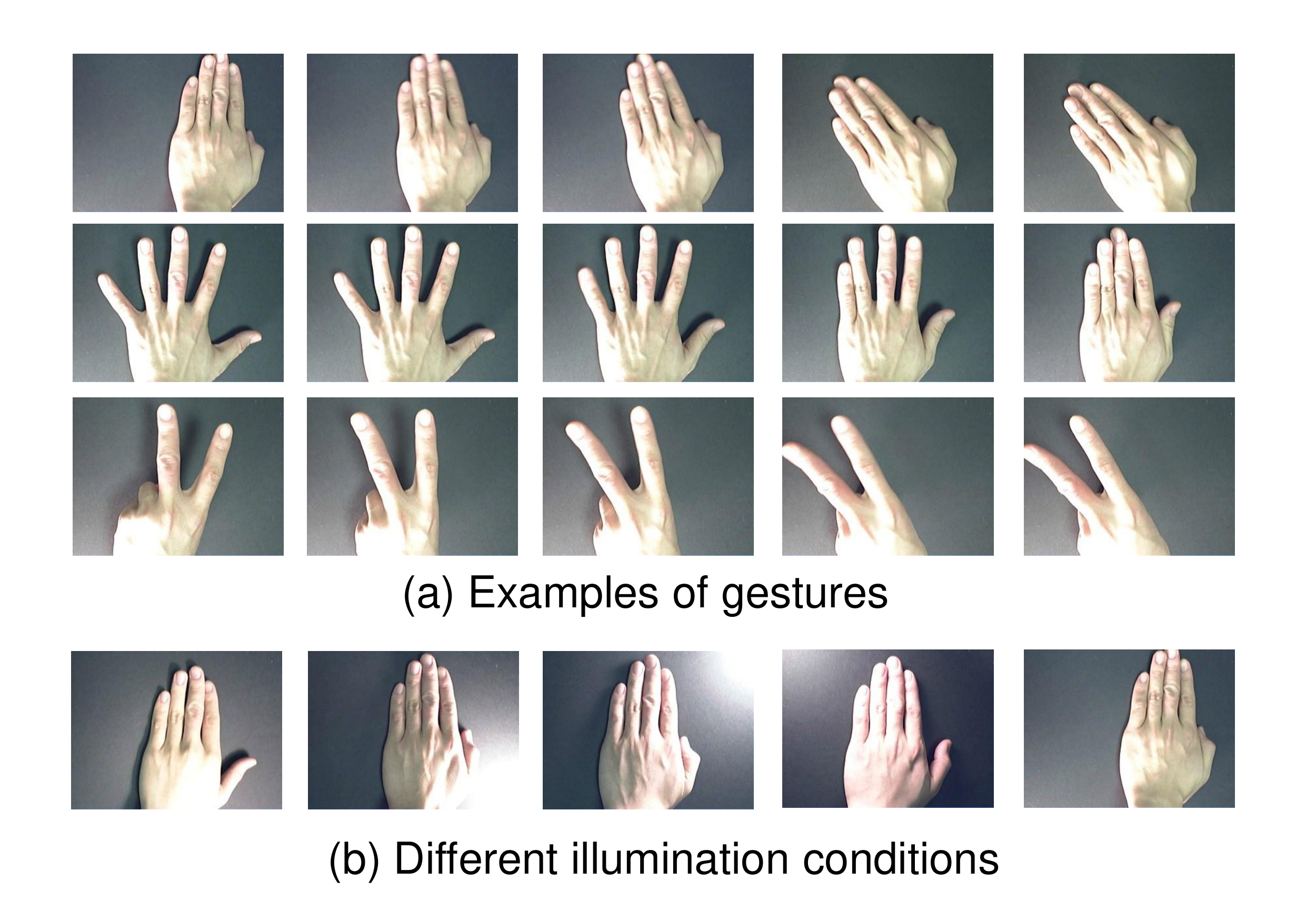}
\end{tabular}
\caption{(a) shows three examples of gestures in the Cambridge hand-gesture database. (b) shows the five different illumination conditions in the database. } 
\label{fig:cambridge_ges}
\end{center}
\end{figure}

Hand gesture recognition using videos is an important research area since gesture is a natural way of communication and expressing intentions. 
People use gestures to depict sign language for deaf, convey messages in loud environment and to interface with computers. Therefore, an accurate and automated  gesture recognition system would broadly enhance human-computer interaction and enrich our daily lives. In this section, we are interested in applying our framework in video-base (dynamic) hand gesture recognition. 
We use the {\it Cambridge hand-gesture dataset} \cite{Kim2007} which has 900 video sequences with nine 
different hand gestures: 100 video sequences for each gesture. The nine gestures
result from 3 primitive hand shapes and 3 primitive motions, and as collected under different 
illumination conditions. Some example gestures are shown in Fig. \ref{fig:cambridge_ges}. The gestures are imaged
under five different illuminations, labeled as Set1, Set2, $\dots$, Set5.

In addition to the illumination variability, the main challenge here comes
from the fact that hands in this database are not well aligned, e.g. the proportion of a hand in an image and the location of the hand 
are different in different video sequences.  To reduce these effects we evenly split one image into four quadrants (upper-left, upper-right, 
bottom-left, bottom-right) with some overlaps.   Each of the four quadrants is represented by a sequence of covariance matrices (on the SPDMs manifold $\tilde{\P}$).
In this experiment, we use HOG features \cite{DT05} to form a covariance matrix per image 
quadrant as follows. We use $2\times2$ blocks of $8 \times 8$ pixel cells with $7$ histogram channels to form HOG features. Those HOG features are 
then used to generate $7\times7$ covariance matrix for each quadrant of each frame. Thus, 
our representation of a video is now given by $t \mapsto \alpha(t) \in \tilde{\P}^4$.

Since we have split each hand gesture into four dynamic parts, the 
total distance between any two hand gestures is a composite of 
four corresponding distances. For each corresponding dynamic part, e.g. the upper-left part,  we first the 
parts across videos (using Algorithm \ref{algo:pairwise}) individually 
and then compare them using the metric $d_q$, denoted by $d_{upl}$. The final distance is obtained using an weighted average of the four parts: $d = \lambda_1 d_{upl} + \lambda_2 d_{upr} + \lambda_3 d_{downl} + \lambda_4 d_{downr}$ and $\sum_{i=1}^4\lambda_i  = 1$. For each illumination set, we use different weights, and these weights are trained using randomly selected half of the data ($90$ video sequences) in that set, and the other half of the data are used for the testing. Table \ref{tab:comphand} shows our results using the nearest neighbor classifier on all five sets. 
One can see that after the alignment, the recognition rate has significant improvement on every set. Also, we have reported the state-of-art results on this database \cite{Lui2012,Lui2010}. One can see that our method outperforms these methods.

\begin{table} \caption{Recognition results on the Cambridge Hand-Gesture dataset}
\label{tab:comphand}
\begin{center}
\begin{tabular}{|c|c| ccccc|}
\hline
\multicolumn{2}{|c|}{Method}& Set1 & Set2 & Set3 & Set4 & Set5 \\
\hline
\multicolumn{2}{|c|}{TCCA \cite{4547427}}&81\% & 81\% & 78\% & 86 \%  & -\\
\hline
\multicolumn{2}{|c|}{RLPP \cite{Harandi2012}}&86\% & 86\% & 85\% & 88 \% & -\\
\hline
\multicolumn{2}{|c|}{PM 1-NN \cite{Lui2010}}&89\% & 86\% & 89\% & 87 \% & -\\
\hline
\multicolumn{2}{|c|}{PMLSR  \cite{Lui2012}}&93\% & 89\% & 91\% & 94 \% & -\\
\hline
\multirow{2}{*}{Our} & before alignment & 94\% & 91\% & 90\% & 88\% &  77\%\\
&{\bf after alignment} & {\bf 99}\% & {\bf 97}\% & {\bf 97}\% & {\bf 96}\% &{\bf 98}\%\\
\hline
\multicolumn{2}{|c|}{Improvement} & {\bf 5}\% & {\bf 6}\% & {\bf 7}\% & {\bf 8}\% & {\bf 21}\% \\
\hline
\end{tabular}
\end{center}
\end{table}

\subsection{Visual Speech Recognition} \label{sec:vsr}
This application is concerned with visual speech recognition (VSR), or {\it lip-reading},  
using close-up videos of human facial movements. Speech recognition is important because it allows computers to interpret human speech and take appropriate actions. It also has applications in biometric security, human-machine interaction, manufacturing and so on. Speech recognition is 
performed through multiple modalities - the common speech data consists of both audio and visual components. In the case the audio information is either not available or it is corrupted by noise, it becomes important to understand the speech using the visual data only. This motivates the need for VSR. 

\begin{figure}
\begin{center}
\begin{tabular}{c}
\includegraphics[height=0.45in]{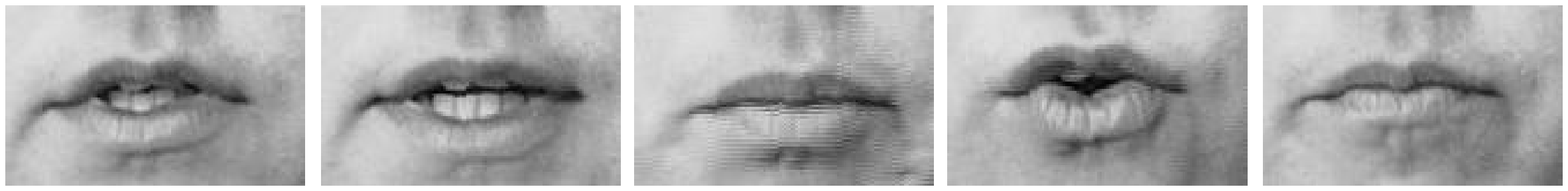}\\
\includegraphics[height=0.45in]{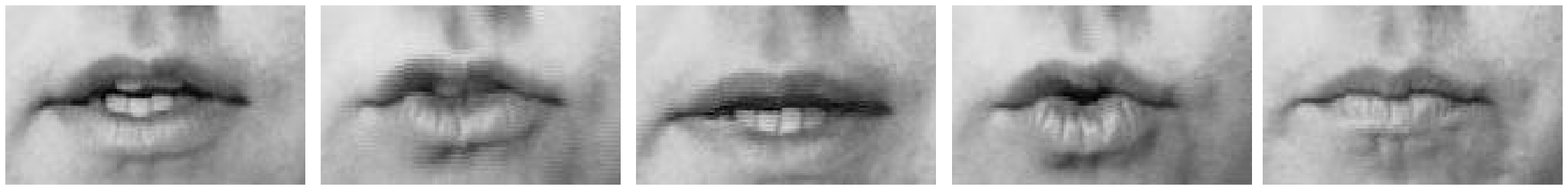}\\
\includegraphics[height=0.46in]{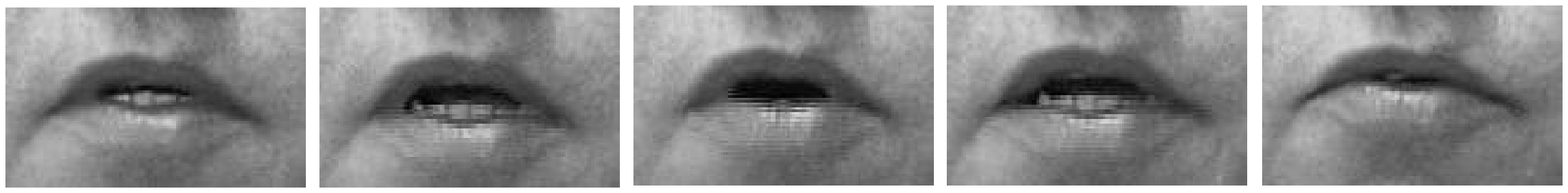}\\
\includegraphics[height=0.425in]{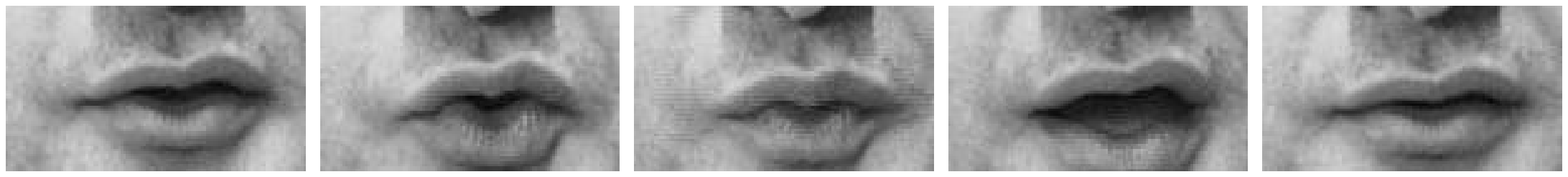}\\
\end{tabular}
\caption{Examples of down sampled video sequences in OuluVS dataset. The first and second row show one person's two speech samples of the phrase ``Nice to meet you''; the third and fourth row show the phrases ``How are you'' and ``Good bye'' uttered by different persons.   } 
\label{fig:ouluvsexamples}
\end{center}
\end{figure}

The process of visual speech recognition is to understand the words uttered by speakers, derived from the visual cues. Movements of the tongue, lips, jaw and other speech related articulators are involved to make sound. To represent a video of such dynamic process, we extract a covariance matrix for each frame.
Now each video becomes a parameterized trajectory  on the space of $\tilde{\P}$ and we can utilize the proposed framework to analysis these trajectories.
In the experiments reported here, we utilize the commonly used OuluVS dataset \cite{Zhao:2009} which includes 20 speakers, each uttering 10 everyday greetings five times:  { \it Hello, Excuse me, I am sorry, Thank you, Good bye, See you, Nice to meet you, You are welcome, How are you, Have a good time.} Thus, totally the database has 1000 videos; all the image sequences are segmented, having the mouth region determined by manually labeled eye positions in each frame \cite{Zhao:2007}. Some examples of the segmented mouth images are shown in Fig. \ref{fig:ouluvsexamples}. 

\begin{figure}
\begin{center}
\begin{tabular}{cc}
\includegraphics[height=1.6in]{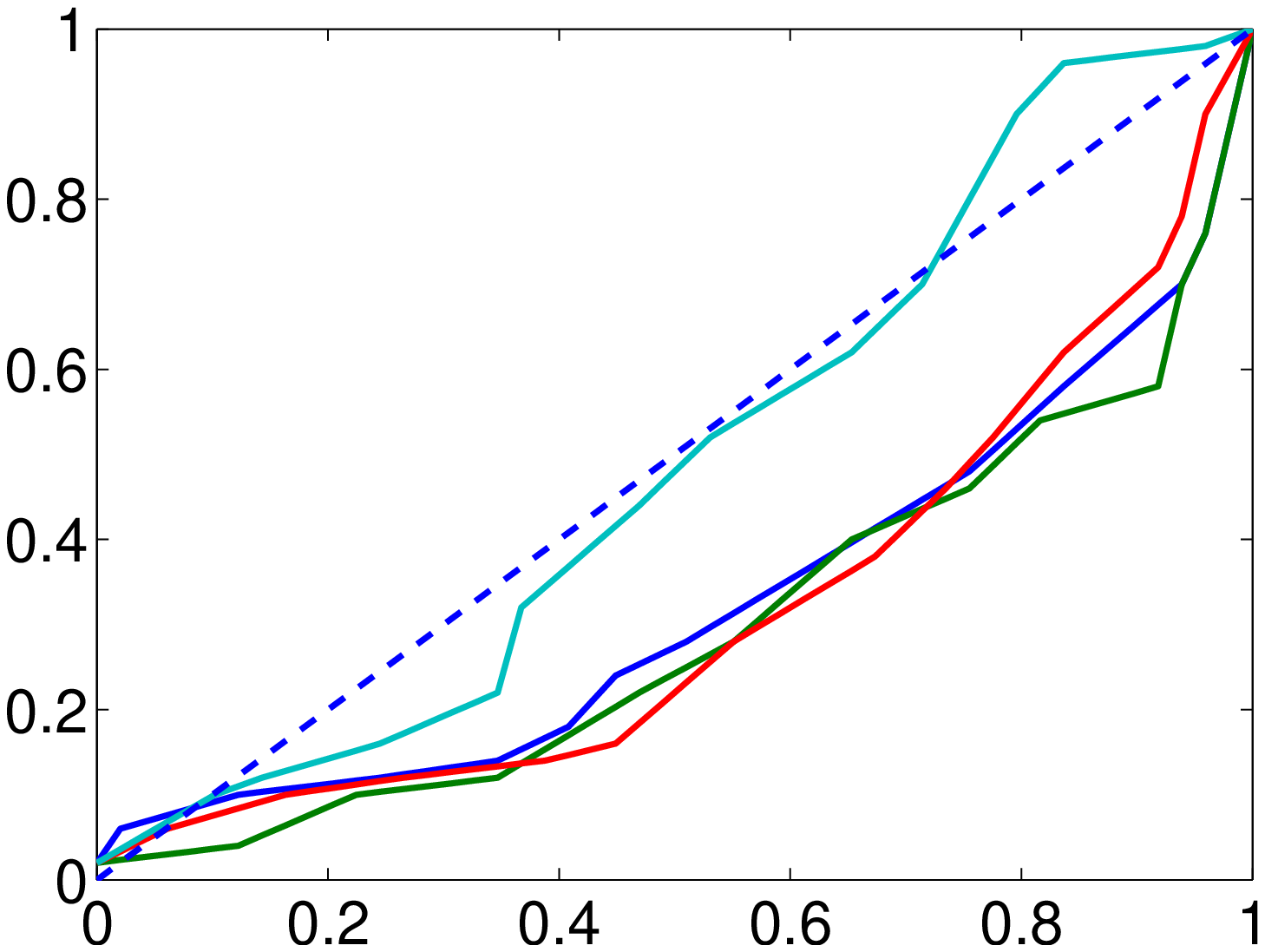}&
\includegraphics[height=1.6in]{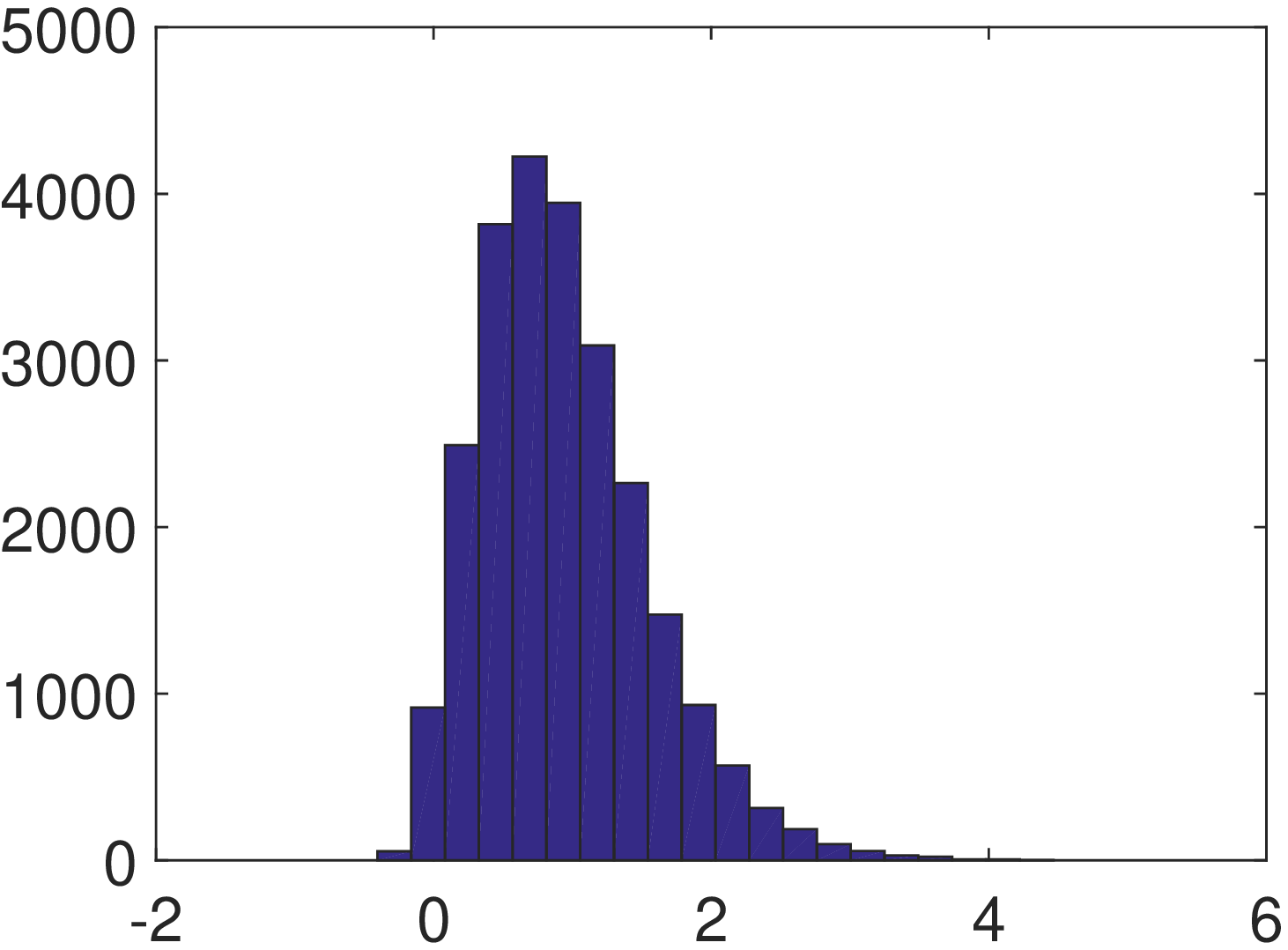}\\
(a) $\gamma^*$ & (b) Histogram of $(d_c - d_q)$'s \\
\end{tabular}
\caption{(a) shows the optimal $\gamma$'s obtained to align one video of phrase (``excuse me'') to the other four videos of the same phrase spoken by same person. (b) shows we show the histogram of $(d_c-d_q)$'s (differences between distances before and after alignment). } 
\label{fig:ouluvsanalysis}
\end{center}
\end{figure}

For each video frame $I$, we extract a $7 \times 7$ covariance matrix representing 
seven features: $\{x,y,I(x,y),$ $ |\frac{\partial I}{\partial x}|, |\frac{\partial I}{\partial y}|, |\frac{\partial^2 I}{\partial^2 x}|, |\frac{\partial^2 I}{\partial^2 y}| \}$. 
The resulting trajectories in $\tilde{\P}$ are aligned using Algorithm \ref{algo:pairwise} and compared using distance $d_q$ defined in Eqn. \ref{geodistq}.  In Fig. \ref{fig:ouluvsanalysis} (a), we show some optimal $\gamma$'s obtained to align one video of phrase (``excuse me'') to other videos of the same phrase spoken by the same person. One can see that 
there exist temporal differences in the original videos and they need to be aligned before further analysis. In (b), we show the histogram of $(d_c-d_q)$'s (differences between distances before and after alignment). In this case, each person has $50$ videos, and we can calculate $(50\times49)/2$ pairwise distances before and after alignment, and their differences. For all $20$ persons in this dataset, we have $20 \times (50\times49)/2  = 24500$ such differences. From the histogram of these differences, one can see that after the alignment, the distances ($d_q$'s) consistently become smaller. Note that one can choose other features to obtain a better representation perhaps, but the main point here is the improvement in temporal alignment and reduced distances between trajectories.  

To compare classification performance with previous methods \cite{Zhao:2009,Su_2014_CVPR}, we perform the experiment on a subset of the whole dataset, which contains $800$ video sequences by removing some short videos due to the restriction of the method in \cite{Zhao:2009}. (Although our method do not have this restriction, to compare fairly, we perform the experiment in the same subset). 
Then, we perform the Speaker-Dependent Test (SDT, see \cite{Zhao:2009} for details) on this subset. The recognition rate is calculated based on Nearest Neighbor (NN) classifier. Table \ref{tab:comp} shows the average first nearest neighbor (1NN) classification rate of our method and previous methods. One can see that even with a simple classifier, our method has the classification rate of $78.6\%$, which is $8.1\%$ better than \cite{Su_2014_CVPR}'s. This indicates the advantages of proposed intrinsic method compared with the one in \cite{Su_2014_CVPR}. One can also see that, there are $37.6\%$ percent of improvement after the alignment (registration), which demonstrates the importance of removing temporal difference in comparing of the dynamic systems in computer vision.  
Several other papers have reported higher classification rates on this dataset by they generally use
advanced classier (e.g. SVM, leave-one-out cross valuation), additional information (e.g. audio, image depth) and machine learning techniques \cite{Pei:2013,5995345,5597428}. Thus, their results are not directly comparable with our results. 

\begin{table} \caption{Comparison of SDT performance on OuluVS database}
\label{tab:comp}
\begin{center}
\begin{tabular}{|c|c|c|}
\hline
\multicolumn{2}{|c|}{Method}& 1NN Rate \\
\hline
\multicolumn{2}{|c|}{Zhao et al. \cite{Zhao:2009}} & 70.2\% \\
\hline
\multirow{2}{*}{Su et al. \cite{Su_2014_CVPR}} & before alignment & 33.8\% \\
&after alignment & 70.5\% \\
\hline
\multirow{2}{*}{Our method} & before alignment & 41.0\% \\
&{\bf after alignment} & {\bf 78.6\%} \\
\hline
\end{tabular}
\end{center}
\end{table}

\section{Conclusion}
In summary, we have proposed metric-based approach  for simultaneous alignment and comparisons of trajectories
on $\tilde{\P}$, the Riemannian manifold of covariance matrices (SPDMs). 
In order to facilitate our analysis, we impose a Riemannian structure on this manifold, induced by the action of
$SO(n)$ and $SL(n)$, resulting in explicit expressions for geometric quantities such as parallel transport and 
Riemannian curvature tensor. 
Returning to the trajectories, the basic idea is to represent each trajectory by 
a starting point $\tilde{P} \in \tilde{\P}$ and a TSRVF which is a curve in the tangent space $\T_{\tilde{P}}(\tilde{\P})$. The metric for comparing
these elements is a composed of: (a) the length of the path between the starting points and (b) the distortion introduced in 
parallel translation TSRVFs along that path. The search for optimal path, or a geodesic, is based on a shooting method,
that in itself uses geodesic equations for computing the exponential map. Using a numerical implementation of the 
exponential map, we derive numerical solutions for pairwise alignment of  covariance trajectories and to quantify 
their differences using a rate-invariant distance. We have applied this framework to covariance tracking in video data, 
with applications to hand-gesture recognition and visual-speech recognition, and obtain state of the art results in 
each case.

\bibliography{bibfile}
\bibliographystyle{IEEEtran}

\newpage
\appendix
\section{Riemannian Structure on $SL(n)$ and Its Quotient Space}
\label{sec:geom}
We start with $SL(n)$, the set of all $n \times n$ with unit determinant. 
For the identity matrix $I \in SL(n)$,
the tangent space at identity is given by  $\mathfrak{sl}(n) \equiv \T_I(SL(n)) =  \{A|\text{tr}(A)=0\}$, the space of all $n \times n$ matrices with trace zeros. 
The tangent space at any other point $G \in SL(n) $, $\T_G(SL(n))$, is given by $\{GA|A \in  \mathfrak{sl}(n)\}$.
Each left-invariant vector field on $SL(n)$ is determined by its value at $I$: $A(G) = GA(I)$, where $G \in SL(n)$ and $A(I) \in  \mathfrak{sl}(n)$.  
Note that any group naturally acts on itself and we will use this fact for $SL(n)$.

The lie algebra $\mathfrak{sl}(n)$ can be expressed as the direct sum of two vector subspaces: $\mathfrak{sl}(n) = \mathfrak{so}(n) \oplus \mathfrak{p}(n)$, 
where $\mathfrak{so}(n) = \{A \in  \mathfrak{sl}(n)| A^t = -A\}$, and $ \mathfrak{p}(n) = \{A \in  \mathfrak{sl}(n)| A^t = A\}$.
The tangent space at any arbitrary point $G \in SL(n)$ also has a similar decomposition: 
for any $B \in {\cal T}_G(SL(n))$, we have $B = G B_s + G B_p$ such that $B_s \in \mathfrak{so}(n)$ and 
$B_p \in \mathfrak{p}(n)$. \\

\noindent {\bf Riemannian metric on $SL(n)$}: 
Now we define the Riemannian metric on $SL(n)$ that will later be used for inducing a Riemannian 
structure on $\P$. For $B,C \in \mathfrak{sl}(n)$, we define a metric as $\inner{B}{C}_I = \text{tr}(BC^t)$. 
At any other point $G \in SL(n)$, the metric is calculated by pulling back the tangent 
vectors $B^\prime,C^\prime \in \mathcal{T}_G(SL(n))$ to $ \mathfrak{sl}(n)$, 
by multiplying $G^{-1}$ on the left, i.e.  $G^{-1}B^\prime, G^{-1}C^\prime$. Thus, the inner product  (or the Riemannian metric) is 
given by: 
\begin{equation}
\inner{B^\prime}{C^\prime}_G = \text{tr}((G^{-1}B^\prime) (G^{-1}C^\prime)^t)\ \ ,
\label{eqn:metric-invariance-sln}
\end{equation}
where $\ G \in SL(n),\ \ B^\prime, C^\prime \in \mathcal{T}_G(SL(n))$. This pullback operation ensures that this
metric is invariant to the left action of $SL(n)$, i.e. $\inner{B}{C}_I = \inner{GB}{GC}_G $.  In fact, the difference 
in this framework to the Riemannian metric used in \cite{Pennec:2006} comes from the mapping used for pullback. 
The mapping used in that paper is not left invariant. \\

\noindent {\bf Geodesic paths on $SL(n)$}: 
 An important consequence of the $SL(n)$-invariance (Eqn. \ref{eqn:metric-invariance-sln}) is that we can 
simplify some calculations by transforming our problems appropriately. For instance, if we need to compute a 
geodesic between two arbitrary elements $G_1, G_2 \in SL(n)$, then we can first solve the problem of 
computing the geodesic,  between $I$ and $G_{12}$, where $G_{12} = G_1^{-1} G_2$, and then 
multiply on the left by $G_1$. The geodesic between $I$ and $G_{12}$ is given by
$t \mapsto e^{A_{12}^t}e^{(A_{12} - A_{12}^t)} $, according to \cite{andruchow:2011}, where $ A_{12} = \argmin_{A \in \mathfrak{gl}(n)}\|e^{A^t}e^{(A - A^t)} -G_{12} \|_F$, and $\|\cdot\|_F$ indicates the Frobenius norm.  
 In fact, if $G_{12}$ is symmetric and positive definite, then $A_{12} = \text{logm}(G_{12}) \in \mathfrak{p}(n)$ and the geodesic has the simple expression $t \mapsto  e^{tA_{12}}$, so that the desired geodesic between $G_1$ and $G_2$ is $t \mapsto G_1 e^{t A_{12}}$. 
The geodesic distance between the two points is given by: 
\begin{equation}
d_{SL(n)}(G_1, G_2) =d_{SL(n)}(I, G_{12}) =  \| A_{12}\|\ .
\label{eqn:geod-dist-GLn}
\end{equation}


\noindent {\bf Riemannian curvature tensor on $SL(n)$}: 
Let $X$, $Y$ and $Z$ be three tangent vectors at a point $G \in SL(n)$, and we want to compute the 
Riemannian curvature tensor $R(X,Y)(Z)$. While the general form for $R(X,Y)(Z)$ is complicated \cite{andruchow:2011}, if $X = GA$, $Y = GB$ and $Z = GC$ where $A,B$ and $C$ are elements of $\mathfrak{p}(n)$, 
 then the tensor is given by: 
$$
R(X,Y)(Z) = -\frac{1}{4}[[X,Y],Z] = -\frac{1}{4}G[[A,B],C]\ ,
$$
where $[A,B] = AB - BA$. (This use of square brackets is called {\it Lie bracket} and should be distinguished from 
the use of square brackets to denote equivalence classes.)
 
Now we consider a quotient space of $SL(n)$ and, using the theory of Riemannian submersion, inherit some of these
formulas to this quotient space. 
Define the right action of $SO(n)$ on $SL(n)$ according to:
 $$
 SL(n) \times SO(n) \rightarrow SL(n),\ \ \mbox{given by}\ \ (G*S) = GS\ .
 $$
An orbit under this action is given by $[G] = \{ GS | S \in SO(n)\}$. The set
 of these orbits forms the quotient space $SL(n)/SO(n) = \{ [G]  | G \in SL(n)\}$. Since $SO(n)$ is a closed subgroup of $SL(n)$, 
 the quotient space is a manifold. 
 
 Using the Riemannian metric on $SL(n)$, it is easy to specify the tangent bundle of the quotient space. 
The tangent space $\T_{[I]}(SL(n)/SO(n))$ is simply the subspace of 
 $\T_I(SL(n))$ which is orthogonal ${\cal T}_I(SO(n))$,  the space tangent to the orbit at $I$. It is well known that  
 $\T_I(SO(n)) \equiv  \mathfrak{so}(n)$. 
The subspace of  $\mathfrak{sl}(n)$ which is orthogonal to $\mathfrak{so}(n)$ is $\mathfrak{p}(n)$. So, we have 
 $\T_{[I]}(SL(n)/SO(n)) = \mathfrak{p}(n)$.
For any arbitrary $G \in SL(n)$, the vector space tangent to the quotient space at $[G]$, 
$\T_{[G]}(SL(n)/SO(n))$ can be identified with the set $\{ \tilde{G} B| B \in \mathfrak{p}(n),\ \mbox{for any}\  \tilde{G} \in [G]\}$. 
 
 \noindent {\bf Riemannian metric on $SL(n)/SO(n)$}:
Now that we have the tangent bundle of the quotient space, we 
can define a Riemannian metric on this quotient space by inducing it from the larger space $SL(n)$. 
This is possible since the action of $SO(n)$ is by isometries under that metric (and the fact that $SO(n)$ is a closed
set). 
By isometry we mean that  $\inner{AS}{BS}_{GS} = \inner{A}{B}_G$, for 
any $S \in SO(n)$ and $A,B \in \mathcal{T}_G(SL(n))$. Therefore, we can induce this metric from $SL(n)$ to the quotient space
$SL(n)/SO(n)$. For any two vectors $\tilde{A}, \tilde{B} \in \T_{[G]}(SL(n)/SO(n))$, with the above identification, 
we define the metric as: 
$\inner{\tilde{A}}{\tilde{B}}_{[G]}  = \text{tr}((\tilde{G}^{-1}\tilde{A})(\tilde{G}^{-1}\tilde{B})^t)$, for any 
$\tilde{G} \in [G]$.
Note that $\tilde{G}^{-1}\tilde{A}, \tilde{G}^{-1}\tilde{B} \in  \mathfrak{p}(n)$.

\noindent {\bf Geodesic paths on $SL(n)/SO(n)$}:
The geodesics in the quotient space $SL(n)/SO(n)$ 
can be expressed using those geodesics in the larger space, $SL(n)$, that
are perpendicular to every orbit they meet. Therefore, a geodesic between the points $[G_1]$ and 
$[G_2]$ in $SL(n)/SO(n)$ is given by $t \mapsto [G_1 e^{t A_{12}}]$, where $A_{12} \in \mathfrak{p}(n)$ such 
that $e^{A_{12}} \in [G_1^{-1} G_2]$. 
The last part means that there exists an $S_{12} \in SO(n)$ such that 
$e^{A_{12}} = G_1^{-1}G_2 S_{12}$. 
The geodesic distance between the two orbits is given by: 
\begin{equation}
d_{SL(n)/SO(n)}([G_1], [G_2]) =d_{SL(n)/SO(n)}([I], [G_{12}]) =  \| A_{12}\|\ .
\label{eqn:dist-quotient}
\end{equation}

\noindent {\bf Parallel transport of tangent vectors along geodesics on $SL(n)/SO(n)$}:
Let $X \in \T_{[G]}(SL(n)/SO(n))$ be a tangent vector that needs to be translated along a geodesic path 
given by $t \mapsto  [Ge^{t A}]$, where $A \in \mathfrak{p}(n)$. Let $B \in \mathfrak{p}(n)$ such that $X$ is identified with $\tilde{G} B$, where $\tilde{G} \in [G]$. 
Then, the parallel transport of 
$X$ along the geodesic is identified with the vector field $t \mapsto \tilde{G} e^{t A} B$ along $\tilde{G}e^{tA}$ on $SL(n)$. 

\noindent {\bf Riemannian curvature tensor on $SL(n)/SO(n)$}:
Let $X$, $Y$ and $Z$ be three tangent vectors at a point $[G] \in SL(n)/SO(n)$, and we want to compute the 
Riemannian curvature tensor $R(X,Y)(Z)$ on the quotient space. Let $A$, $B$ and $C$ be elements of $\mathfrak{p}(n)$ such that
$X = \tilde{G}A$, $Y = \tilde{G}B$ and $Z = \tilde{G}C$, where one can use any $\tilde{G} \in [G]$ for this 
purpose. Then, the tensor is given by: 
$$
R(X,Y)(Z) = -[[X,Y],Z] = -\tilde{G}[[A,B],C]\ ,
$$
where $[A,B] = AB - BA$ as earlier.

\end{document}